\documentclass[11pt]{article}
\usepackage{rotating,amsmath,amssymb,amsfonts,amsthm}
\usepackage{epsfig,mathrsfs,color,graphics,comment}
\usepackage{multirow}
\usepackage{amsmath}
\usepackage{mathtools}
\def\boxit#1{\vbox{\hrule\hbox{\vrule\kern6pt \vbox{\kern6pt#1\kern5pt}
\kern6pt\vrule}\hrule}}

\usepackage{geometry}
\usepackage[textsize=tiny]{todonotes}

\usepackage{rotating,amsmath,amssymb,amsfonts,amsthm}
\usepackage{epsfig,mathrsfs,natbib,color}
\usepackage{graphicx,graphics}
\usepackage{siunitx,lmodern}
\usepackage{array,wrapfig,subfigure,caption}
\usepackage{hyperref}
\usepackage{booktabs,adjustbox,arydshln,multirow}
\usepackage[utf8]{inputenc}  
\usepackage[T1]{fontenc}     
\usepackage{url}             
\usepackage{nicefrac}   
\usepackage{microtype}    
\usepackage{natbib}
 \usepackage{algorithm}
 \usepackage{algpseudocode}
\usepackage{tikz}

\DeclarePairedDelimiter\ceil{\lceil}{\rceil}

\newcommand{\bK}{{\boldsymbol K}}

\newcommand{\bQ}{{\boldsymbol Q}}

\newcommand{\bU}{{\boldsymbol U}}
\newcommand{\bV}{{\boldsymbol V}}

\newcommand{\bZ}{{\boldsymbol Z}}
\newcommand{\bS}{{\boldsymbol S}}

\newcommand{\ba}{{\boldsymbol a}}

\newcommand{\bu}{{\boldsymbol u}}

\newcommand{\bx}{{\boldsymbol x}}
\newcommand{\by}{{\boldsymbol y}}
\newcommand{\bz}{{\boldsymbol z}}

\newcommand{\bbeta}{{\boldsymbol \beta}}
\newcommand{\bgamma}{{\boldsymbol \gamma}}
\newcommand{\btheta}{{\boldsymbol \theta}}
\newcommand{\bepsilon}{{\boldsymbol \epsilon}}

\begin{document}


 \title{Magnitude Pruning of Large Pretrained Transformer Models with a Mixture Gaussian Prior}
 
\author{Mingxuan Zhang, Yan Sun, and Faming Liang\thanks{ 
 M. Zhang (email: zhan3692@purdue.edu) and Y. Sun (email: sun748@purdue.edu) 
  are Graduate Students, and  
F. Liang (email: fmliang@purdue.edu) is Distinguished Professor,  Department of Statistics, Purdue University, West Lafayette, IN 47907.
 } 
 }

\maketitle
 
 \begin{abstract}
   Large pretrained transformer models have revolutionized modern AI applications with their state-of-the-art performance in natural language processing (NLP). However, their substantial parameter count poses 
   challenges for real-world deployment. To address this, researchers 
   often reduce model size by pruning parameters based on their magnitude 
   or sensitivity. 
  Previous research has demonstrated the limitations of magnitude pruning, especially in the context of transfer learning for modern NLP tasks. 
  In this paper, we introduce a new magnitude-based pruning algorithm called mixture Gaussian prior pruning (MGPP), which employs a mixture Gaussian prior for regularization. MGPP prunes non-expressive weights under the guidance of the mixture Gaussian prior, aiming to retain the model's expressive capability. Extensive evaluations across various NLP tasks, including natural language understanding, question answering, and natural language generation, demonstrate the superiority of MGPP over existing pruning methods, particularly in high sparsity settings. Additionally, 
  we  provide a theoretical justification for the consistency of the sparse transformer, shedding light on the effectiveness of the proposed pruning method. 

\end{abstract}

\section{Introduction}

Large pretrained transformer models have emerged as powerful tools for a variety of downstream natural language processing tasks, from natural language generation to question answering \citep{radford2019language, brown2020language}. These pretrained models have grown exponentially in size, often comprising hundreds of millions, or even billions, of parameters \citep{devlin2019bert, he2021debertav3, lewis2019bart, touvron2023llama}. While their capabilities are undeniably impressive, the computational and storage requirements for such large models are becoming increasingly prohibitive \citep{strubell2020energy}.

Score-based pruning, a technique that involves removal of non-expressive parameters based on their importance score rankings, presents a promising avenue for model compression. It has the potential to significantly reduce model size with minimal impact on performance.
Based on the definition of pruning scores, the pruning methods can be 
classified into distinct categories, such as magnitude-based (zeroth-order) pruning methods \citep{han2015deep, han2015learning, zhu2017prune, louizos2017learning, wang2020neural} and sensitivity-based (higher-order) pruning methods \citep{molchanov2019importance, ding2019global, sanh2020movement, liang2021super, zhang2022platon, kurtic2022optimal, li2023losparse}. On the other hand, if the classification is made based on the strategies employed, the methods fall into categories such as one-shot pruning \citep{lee2018snip, frankle2018lottery, chen2020lottery, liang2021super, zafrir2021prune} and iterative pruning \citep{han2015deep, zhu2017prune, louizos2017learning, sanh2020movement, zhang2022platon, li2023losparse}.

It has long been argued and experimentally demonstrated that magnitude-based pruning methods struggle to retain expressive parameters, particularly in high-sparsity settings. Furthermore, when it comes to transfer learning with large pretrained models, which are now the benchmark for state-of-the-art downstream NLP tasks, their effectiveness is reduced. As a result, models pruned using magnitude-based methods often exhibit diminished generalization performance \citep{sanh2020movement}.

Recently, Bayesian sparse deep learning has made significant progress through a series of works \citep{liang2018bayesian, sun2022learning, sun2021consistent, sun2021sparse, zhang2023sparse}, demonstrating its potential in deep learning for both statistical inference and model sparsification. By adopting the mixture Gaussian prior (MGP) for the 
parameters of the neural network, 
they developed a magnitude-based one-shot pruning method and 
achieved state-of-the-art performance in pruning both convolutional neural networks \citep{sun2021consistent, sun2021sparse} and recurrent neural networks \citep{zhang2023sparse}. These early experimental results on small-scale models have once again sparked hope for magnitude-based pruning methods. However, their methods have not yet been evaluated on large transformer models across different tasks and datasets. Moreover, as we will discuss in Section \ref{main:sparse_deep_learning}, several key challenges prevent us from directly adopting their methods for pruning larger models.

In this work, we introduce MGPP, a magnitude-based iterative pruning algorithm that is both simple and effective. To validate its performance, we conducted extensive experiments across three key downstream tasks: natural language understanding, question answering, and natural language generation. Our evaluations span three types of pretrained transformer-based language models, DeBERTaV3\textsubscript{base} \citep{he2021debertav3}, BERT\textsubscript{base} \citep{devlin2019bert}, and BART\textsubscript{large} \citep{lewis2019bart}. Our results indicate that when guided by an appropriate prior, magnitude-based methods can outperform existing state-of-the-art pruning methods. Additionally, we provide a loose justification for the consistency of the sparse transformer, shedding 
light on its effectiveness. 
 
The remaining part of the paper is organized as follows. Section 2 provides preliminary descriptions for related concepts and methods 
in the literature. Section 3 describes the proposed method and justifies 
its validity. 
Section 4 reports numerical experiments. Section 5 concludes the paper 
with a brief discussion. 

\section{Preliminaries}

\subsection{Pruning Scores}
\label{main:pruning_scores}


An essential component of effective pruning is accurately identifying 
non-expressive parameters through their importance score rankings. Consider a model defined by a set of parameters $\btheta = (\theta_1, \ldots, \theta_d)^T \in \mathbb{R}^d$, each associated with an importance score.
 Let $\bS = (S_1, \ldots, S_d)^T \in \mathbb{R}^d$ denote the corresponding 
  score vector. Score-based pruning methods eliminate parameters based on these scores, with parameters assigned lower scores being prioritized for removal.  As outlined in the Introduction, score-based pruning methods fall into two primary categories, namely, magnitude-based methods and sensitivity-based methods.

\paragraph{Magnitude-Based (Zeroth-Order) Methods} \citep{zhu2017prune, wang2020neural, chen2020lottery, zafrir2021prune}, which determine the 
 parameters to prune based on their magnitudes.  
For a given parameter $\theta_j$, the score is defined as $S_j = |\theta_j|$. Among these methods, gradual magnitude pruning (GMP), introduced by \citet{zhu2017prune}, is particularly notable for its effectiveness and simplicity. This widely adopted pruning baseline has inspired the development of numerous subsequent methods, see e.g., \cite{chen2020lottery} and \cite{zafrir2021prune}.

\paragraph{Sensitivity-Based Methods} \citep{sanh2020movement, zhang2022platon, kurtic2022optimal, li2023losparse},
which incorporate higher-order information, such as gradients and Hessian, to assess the impact of pruning on the loss function \(\mathcal{L}\).
The first-order pruning methods utilize gradient-based information. 
Notable examples include movement pruning (MvP) \citep{sanh2020movement} 
and PLATON \citep{zhang2022platon}. The former removes model parameters 
that are moving towards zero, and the latter is designed to capture the uncertainty of model parameters' importance scores during the pruning process. In downstream pruning scenarios, particularly for BERT-like language models, PLATON is recognized as state-of-the-art, significantly outperforming other baselines, including MvP.
The second-order pruning methods \citep{lecun1989optimal, singh2020woodfisher, frantar2021m, kurtic2022optimal} utilize Hessian-based information. To circumvent the costly approximation of the inverse Hessian, \citet{singh2020woodfisher} introduced the WoodFisher method, and  \cite{frantar2021m} introduced the M-FAC method. However, \citet{kurtic2022optimal} showed that the WoodFisher method is computationally infeasible at the scale of BERT, and while the M-FAC method 
scales effectively, it yields inferior pruning results. In response, they proposed a general second-order pruning method, 
Optimal BERT Surgeon (oBERT), which achieves state-of-the-art performance 
in upstream pruning scenarios.

While the zeroth-order methods are simple, scalable, and often serve as standard baselines, they are consistently outperformed by higher-order methods, particularly in downstream pruning scenarios and at high sparsity levels. However, as discussed above, the performance gains from utilizing higher-order information come at the cost of additional memory and computational resources. For a model with \(d\) parameters, PLATON requires an extra $O(3d)$ memory to maintain three additional states: the average importance scores between consecutive pruning operations, and the exponential average of both the importance scores and the corresponding upper confidence bound. For the BERT\(_{\text{BASE}}\) model, with \(d = 85\) million parameters, managing these states is feasible. However, scaling to larger models, such as Llama-7b/70b \citep{touvron2023llama}, necessitates approximately an additional 84GB/840GB of GPU memory.

The memory requirement for oBERT is $O(Bd)$, where \(B\) is a hyperparameter representing the width of the diagonal block-wise approximation of the empirical Fisher matrix. For the BERT\(_{\text{BASE}}\) model, setting \(B = 50\) results in an additional memory requirement of about 17GB. This demand is manageable for BERT\(_{\text{BASE}}\) but becomes unscalable for larger models. The runtime complexity of oBERT is $O(mBd)$, where \(m\) denotes the number of gradient outer products used to approximate the Hessian; for the BERT\(_{\text{BASE}}\) model, \(m\) is set to 1024.

\subsection{Pruning Strategies}
\label{main:pruning_strategies}

Pruning strategies can be classified into one-shot pruning and iterative pruning. In one-shot pruning, the sparsity pattern is predetermined using the scores of a fully-trained, dense model. A sparse model is then trained with pruned parameters fixed, a  technique often termed "rewinding." However, choosing which parameters to prune based on a fully-trained model overlooks the complex dynamics of training. As a result, parameters that are expressive may be unfairly eliminated at the early stage of training.

On the other hand, iterative pruning jointly performs training and pruning. The sparsity pattern is dynamically updated, offering the model an opportunity to recover from previous pruning decisions. Additionally, the sparsity level can be gradually increased during training through sparsity schedulers, such as the cubic sparsity scheduler \citep{zhu2017prune, sanh2020movement, zafrir2021prune, zhang2022platon, kurtic2022optimal, li2023losparse} given as follows:
\begin{equation}
\label{cubic_sparse_scheduler}
    v^{(t)} = \begin{cases} 
      0 & t < t_i, \\
      v^{(T)} - v^{(T)}\left(1 - \frac{t-t_i}{t_f-t_i}\right)^3 & t_i \leq t \leq t_f, \\
      v^{(T)} &  t_f < t \leq T,
    \end{cases}
\end{equation}
where $v^{(t)}$ is the sparsity level at the $t$-th training step, increasing from an initial value 0 to a final level $v^{(T)}$ over a period of $T-t_i-t_f$ steps following a warm-up of $t_i$ steps. In practice, instead of performing pruning at every training step before reaching the target sparsity level, one can also choose to prune every \( \Delta t \) steps \citep{zhang2022platon, li2023losparse}.

\subsection{Mixture Gaussian Priors in Bayesian Sparse Deep Learning}
\label{main:sparse_deep_learning}

The mixture Gaussian prior (MGP) has recently attracted significant attention in the field of Bayesian sparse deep learning \citep{sun2021consistent}. Formally, it models each parameter of the network using a mixture Gaussian prior, defined as follows:
\begin{equation} 
    \label{mg}
    \theta_{j} \sim \lambda \cdot \mathcal{N}(0, \sigma^{2}_{1}) + (1-\lambda) \cdot \mathcal{N}(0, \sigma^{2}_{0}),
\end{equation}
where \( \lambda \in (0, 1) \) is the mixture proportion,  \( \sigma^{2}_{0} \) is typically set to a very small value, whereas \( \sigma^{2}_{1} \) is usually assigned a relatively larger value. In what follows, we denote the   prior density function of $\theta_j$ as \( \pi(\theta_j;\lambda, \sigma_0^2, \sigma_1^2) \). Furthermore, we assume that all model parameters are {\it a priori} independent, i.e.,  $\pi(\btheta;\lambda, \sigma_0^2, \sigma_1^2)=\prod_{j=1}^{d}\pi(\theta_j;\lambda, \sigma_0^2, \sigma_1^2)$.  

This particular prior has been shown to offer several theoretical advantages under the Bayesian framework. These include posterior consistency, structure selection consistency, and asymptotic normality of predictions for both i.i.d. data~\citep{sun2021consistent, sun2021sparse} and time-series data~\citep{zhang2023sparse}. These properties make it useful in various applications like variable/model selection, uncertainty quantification, and model sparsification.

Next, we will discuss how this prior is used to prune models. Essentially, the prior serves as a form of regularization, imposing penalty on model parameters. During training, the learning objective becomes 
\begin{equation}
    \label{learning_obj}
    \mathcal{L}(D_n, \btheta) - \dfrac{1}{n}\log(\pi(\btheta;\lambda, \sigma_0^2, \sigma_1^2)), 
\end{equation}
where 
\( n \) denotes the size of the training dataset $D_n$, and 
$\mathcal{L}(D_n, \btheta)$ represents the negative 
log-likelihood function of the deep neural network. 
To facilitate the application of gradient-based optimization algorithms for minimizing (\ref{learning_obj}), we provide the following numerically stable expression of the gradient of log-prior, despite its straightforward derivation:
\begin{equation}
\label{pior_penalty}
\begin{aligned}
       &  \nabla_{\theta_j}\log(\pi(\theta_j;\lambda, \sigma_0^2, \sigma_1^2)) 
       =- \left( \dfrac{\theta_j}{\sigma_0^2} g(\theta_j) + \dfrac{\theta_j}{\sigma_1^2} \left[1-g(\theta_j)\right] \right),
\end{aligned}
\end{equation}
where 
\[
g(\theta_j) = \left( \exp\{ c_2\theta_j^2 + c_1 \} + 1 \right)^{-1},
\]
with $c_1 = \ln(\lambda) - \ln(1-\lambda) + 0.5\ln(\sigma_0^2) - 0.5\ln(\sigma_1^2)$ and  $c_2 = 0.5/\sigma_0^2 - 0.5/\sigma_1^2$. 

\begin{figure*}[htpb]
\centering
\includegraphics[width=\textwidth]{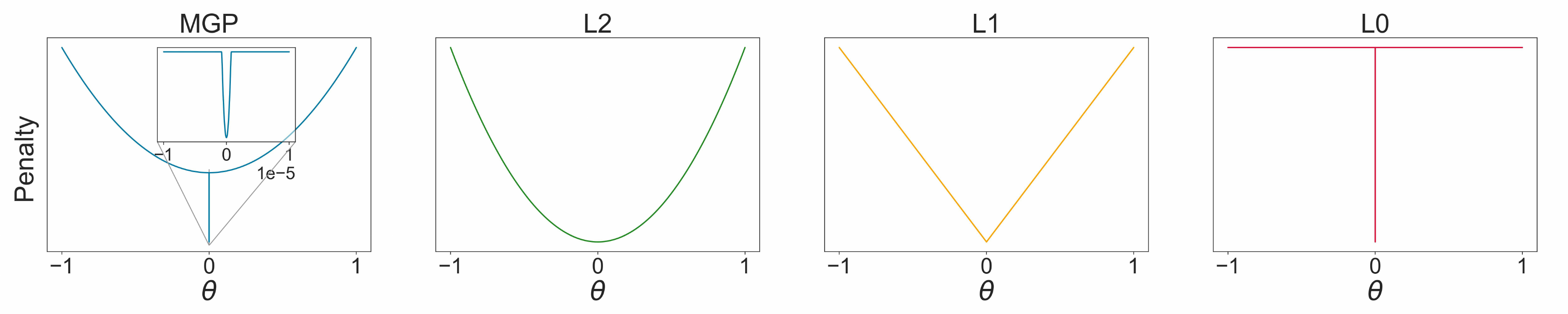}
\caption{Visualization of the penalty functions across various regularization methods: The MGP displayed in the plot corresponds to \( -\log(\pi(\theta;\lambda=1\times10^{-6}, \sigma_0^2=1\times10^{-7}, \sigma_1^2=0.1)) \), where a zoomed-in view for the region near zero is provided. Unlike \( L_0 \) regularization, which is not differentiable and requires the use of gradient estimators \citep{louizos2017learning}, the MGP is differentiable across the entire parameter space.}
\label{fig:diff_regularization}
\end{figure*}

The guidance from the MGP is conveyed through the gradients. The degree of penalty, which serves as the force pushing the model parameters toward zero, can be quantified by the absolute value of the gradient. A comparison between \( L_0 \), \( L_1 \), \( L_2 \), and MGP is presented in Figure~\ref{fig:diff_regularization}.

The MGP acts as a piece-wise \( L_2 \) regularization, imposing different penalties across various regions of the parameter space. On a larger scale, the MGP applies penalties to parameters in a manner similar to \( L_2 \) regularization. In contrast, near zero in the small-scale region, the MGP imposes a more substantial penalty, setting it apart from \( L_2 \) regularization. In Section \ref{appendix:mgps}
of the Appendix, we illustrate and visualize how \( \lambda \), \( \sigma_0^2 \), and \( \sigma_1^2 \) affect the landscape of the MGP.

According to the definitions provided in Sections \ref{main:pruning_scores} and \ref{main:pruning_strategies}, previous methods employing MGP \citep{sun2021consistent, sun2021sparse, sun2022learning, zhang2023sparse} can be categorized as magnitude-based, one-shot pruning methods. These methods train the model using the learning objective specified in Equation (\ref{learning_obj}). Upon convergence, they perform one-shot pruning based on a pruning threshold determined by the values of \(\lambda\), \(\sigma_0^2\), and \(\sigma_1^2\). Then, the pruned model is retrained using only the loss function \(\mathcal{L}\) with the pruned parameters fixed to 0. The detailed algorithm is provided in the Appendix \ref{appendix:pa}.


Their algorithms have set new standards in performance for pruning smaller models like ResNet-20, ResNet-32, and LSTMs. This highlights the effectiveness of MGP. However, translating these successes to larger, transformer-based models introduces challenges due to the many sensitive hyperparameters involved. Additionally, the retraining stage requires further tuning of hyperparameters such as learning rate and batch size, adding another layer of complexity. This is a particular concern given the computational resources required to train larger models.

Besides the aforementioned challenges, achieving a target sparsity level \( v^{(T)} \) adds further complexity. In their approach, \( \sigma_{1}^2 \) and \( \lambda \) are held constant, while \( \sigma_{0}^2 \) is initialized to a large value, denoted as \( (\sigma_{0}^{\text{init}})^2 \). This initial setting is designed to closely align the proportion of pretrained model parameters that fall below the initial pruning threshold with \( v^{(T)} \). A linear scheduler then gradually reduces \( \sigma_{0}^2 \) from \( (\sigma_{0}^{\text{init}})^2 \) to \( (\sigma_{0}^{\text{end}})^2 \). Throughout this prior-annealing (PA) process, the MGP penalty on these parameters increases, effectively driving them toward zero, so that they can be one-shot pruned in the end. However, as explained in Section \ref{main:pruning_strategies}, this pruning strategy may hurt the performance of the sparsified model. We provide experimental evidence in Section \ref{main:ablation_study} to support our arguments.

\section{The MGPP Method}

To overcome these challenges, we introduce the MGPP method, summarized in Algorithm \ref{mgpp_algorithm}. Instead of relying on annealing the MGP to shrink the parameter down to zero, which tends to fix the sparsity pattern too early in the training process, we take a different approach.
We keep the MGP fixed during training and utilize the cubic sparsity scheduler to gradually prune model parameters. When a set of parameters is pruned (i.e., set to zero), they receive gradients only from the loss function in the subsequent training iteration, as the gradients from the MGP become zero. This can serve as a remedy for false selection, thereby overcoming premature pruning of critical parameters.
Parameters with large gradients from the loss are more likely to escape the region with large penalties, giving them a chance to be reconsidered for pruning later.  Conversely, parameters that receive smaller gradients from the loss function will likely remain within the penalized region, making them candidates for future pruning.

Our proposed algorithm introduces only three additional hyperparameters beyond the standard ones: \( \lambda \), \( \sigma_{0}^2 \), and \( \sigma_{1}^2 \). A comprehensive hyperparameter-sensitivity analysis is given in the Appendix \ref{appendix:hs}. Briefly, we find \( \lambda \) to be robust and set it universally to \( 10^{-7} \). Preliminary experiments suggest that, when combined with a sparsity scheduler, it is preferable to set \( \sigma_{0}^2 \) to very small values such as \( 1\times10^{-10} \), in contrast to previous works that often set \( \sigma_{0}^2 \) (specifically, \( (\sigma_{0}^{\text{end}})^2 \)) to larger values, i.e., \( [1\times10^{-7}, 1\times10^{-5}] \). We limit \( \sigma_{0}^2 \) and \( \sigma_{1}^2 \) to the sets \( \{1\times10^{-9}, 1\times10^{-10}\} \) and \( \{0.05, 0.1\} \), respectively. Our approach significantly reduces the computational burden associated with hyperparameter tuning, especially when it comes to training large transformer models. Despite these restrictions, our method outperforms other baselines, as demonstrated in the experimental results section (see Section \ref{main:experiments}).

We also found that gradually incorporating the MGP improves the performance of the sparsified model. This 'prior warm-up' can be seamlessly integrated into the warm-up phase of the sparsity scheduler, denoted by Equation \ref{cubic_sparse_scheduler}. Let \(\eta^{(t)}\) be the prior coefficient at training step \(t\), we have
\begin{equation}
\label{prior_sparse_scheduler}
    v^{(t)}, \eta^{(t)} = \begin{cases} 
      0,\dfrac{t}{t_i} & t < t_i \\
      v^{(T)} - v^{(T)}\left(1 - \frac{t-t_i}{t_f-t_i}\right)^3,1 & t_i \leq t \leq t_f \\
      v^{(T)},1 &  t_f < t \leq T
    \end{cases}
\end{equation}

\begin{algorithm}
\caption{MGPP}
\label{mgpp_algorithm}
\begin{algorithmic}[1]

\State \textbf{Input:} training dataset $D_n$, pretrained model $\btheta^{(0)}$, number of training epochs $E$, mini-batch size $m$, $\lambda$, $\sigma_0^2$, $\sigma_1^2$, $t_i$, $t_f$, $\Delta t$

\State \textbf{Initialize:} $t=1$, $T = \ceil*{En/m}$, optimizer (e.g., AdamW \citep{loshchilov2019decoupled})

\For{epoch from $1$ to $E$}
    \For{each mini-batch \( \mathcal{B} \) sampled from \( D_n \)}
        \State Calculate gradients of the loss through backpropagation
        \begin{equation*}
            \nabla_{\btheta^{(t-1)}}\mathcal{L}(\mathcal{B}, \btheta^{(t-1)})
        \end{equation*}
        \State Calculate $v^{(t)}$ and $\eta^{(t)}$ based on Eq. \ref{prior_sparse_scheduler}
        \State Calculate gradients of MGP based on Eq. \ref{pior_penalty}
        \begin{equation*}
        -\eta^{(t)}\dfrac{1}{n}\nabla_{\btheta^{(t-1)}}\log(\pi(\btheta^{(t-1)};\lambda, \sigma_0^2, \sigma_1^2))
        \end{equation*}
        \State Update $\btheta^{(t-1)} \to \btheta^{(t)}$ by an optimization step
        \State Calculate scores $\bS^{(t)} = (|\theta_1^{(t)}|,\dots,|\theta_d^{(t)}|)$
        \If{$ t \bmod \Delta t = 0 $ or $t > t_f$}
            \begin{equation*}
                \theta^{(t)}_j = 
                    \begin{cases}
                        \theta^{(t)}_j & \text{if $S^{(t)}_j$ in top $v^{(t)}\%$}\\
                        0 & \text{otherwise}
                    \end{cases}
            \end{equation*}
        \EndIf 
        \State Set $t=t+1$
    \EndFor
\EndFor
\end{algorithmic}
\end{algorithm}

In the Appendix \ref{Transformersection}, we provide a loose justification for the parameter estimation consistency of the sparse transformer model with the mixture Gaussian prior, drawing upon the established  
theory from \cite{LiangSLiang2022StoNet} and \cite{Liang2018missing}. 
This justifies the use of the mixture Gaussian prior for sparsifying the transformer model as proposed in the paper, while the MGPP method  proposed above is mainly for 
locating a maximum {\it a posteriori} (MAP) solution for the complex transformer model. Additionally, we note that the parameter estimation
consistency for the sparse transformer model is subject to loss-invariant 
transformations. That is, the model is assumed to be unique up to loss-invariant transformations, e.g., reordering the hidden neurons of the same hidden layer or simultaneously changing the signs or scales of certain connection weights and biases.  The same assumption has often been used in studying theoretical properties of deep neural network models, see e.g., \cite{liang2018bayesian} 
and \cite{sun2021consistent}.

\section{Experiments}
\label{main:experiments}

\subsection{Experimental Setup}

The performance of the final pruned models can be influenced by various factors unrelated to the pruning methodology, including the number of training epochs, the maximum input sequence length, maximum gradient norm, and the number of beams used for evaluation in natural language generation tasks, among others. To control for these variables and ensure a fair comparison with different baselines, we follow the guidelines established in the recent works
\citep{zhang2022platon, kurtic2022optimal, li2023losparse}. We set all these methodology-unrelated factors to match those used in 
\citep{zhang2022platon, kurtic2022optimal, li2023losparse}.
We only tune methodology-related factors, such as \( \lambda \), \( \sigma_1^2 \), and \( \sigma_0^2 \), along with standard hyperparameters like learning rate and batch size, which are also tuned in the baseline methods. Additional details are provided below and in the Appendix \ref{appendix:all_exp}.

We evaluate the proposed method, MGPP, across three downstream NLP tasks: natural language understanding, question answering, and natural language generation, as well as in the upstream pruning scenario. Specifically, we apply MGPP to three pretrained transformer-based language models: \textcolor{black}{DeBERTaV3\textsubscript{base} (180 million parameters), BERT\textsubscript{base} (110 million parameters), and BART\textsubscript{large} (400 million parameters).}

Following the prior works \citep{louizos2017learning, sanh2020movement, zhang2022platon, kurtic2022optimal, li2023losparse}, we prune all weight matrices, except for embeddings, LayerNorm, and the final prediction module. Our implementation is based on the publicly available Hugging Face Transformers library \citep{wolf-etal-2020-transformers}. All performance metrics reported for MGPP are derived from the mean of five independent runs, each using a different random seed.


We compare MGPP with the following baselines:

\begin{itemize}

    
    \item \texttt{Gradual Magnitude Pruning (GMP)} \citep{zhu2017prune} is a simple yet strong magnitude-based iterative pruning baseline, widely recognized  as one of the best magnitude-based pruning methods.

    \item \texttt{Movement Pruning (MvP)} \citep{sanh2020movement} is a sensitivity-based (first-order) iterative pruning method that prunes parameters based on their movement away from zero.
    
    \item \texttt{Iterative pruning (ITP)} \citep{molchanov2019importance} is a sensitivity-based (first-order) iterative pruning method that prunes parameters at each iteration if their importance scores fall below a hard threshold.

    \item \textcolor{black}{\texttt{PLATON} \citep{zhang2022platon} is a sensitivity-based (first-order) iterative pruning method designed to capture the uncertainty of model parameters' importance scores during the pruning process.}
    
    \item \textcolor{black}{\texttt{oBERT} \citep{kurtic2022optimal} is a sensitivity-based (second-order) iterative pruning method that utilizes a diagonal block-wise approximation of the empirical Fisher matrix.}

    \item \texttt{LoSparse} \citep{li2023losparse} is a sensitivity-based (first-order) iterative pruning method for transformer-based language models that integrates low-rank and sparse matrices to prune weight matrices effectively.
    
\end{itemize}

\subsection{Natural Language Understanding}

We assess the pruning performance of MGPP on BERT\textsubscript{base} \citep{devlin2019bert} and DeBERTaV3\textsubscript{base} models \citep{he2021debertav3} by conducting experiments on the General Language Understanding Evaluation (GLUE) benchmark \citep{wang2018glue}, which includes a variety of tasks. Specifically, GLUE features two single-sentence classification tasks, SST-2 \citep{socher2013recursive} and CoLA \citep{warstadt2019neural}, as well as three tasks focused on similarity and paraphrasing: MRPC \citep{dolan2005automatically}, STS-B \citep{cer2017semeval}, and QQP. Additionally, the benchmark includes four natural language inference tasks: MNLI \citep{williams2017broad}, QNLI \citep{rajpurkar2016squad}, RTE \citep{dagan2005pascal}, and WNLI \citep{levesque2012winograd}. In accordance with prior studies, we omit WNLI from our experiments. Additional details regarding the datasets can be found in the Appendix \ref{appendix:nlu}.

A table containing training details, such as learning rate, batch size, the number of training epochs, $\sigma_0^2$, and $\sigma_1^2$ for each dataset, is presented in the Appendix \ref{appendix:nlu}.

The results on the GLUE development set are summarized in Tables \ref{table:glue_debertav3-base} and \ref{table:glue_bert-base}; all baseline results are directly taken from \cite{zhang2022platon, li2023losparse}. MGPP consistently achieves equal or superior performance compared to existing approaches across most datasets and sparsity levels. Notably, as the amount of training data increases, our method performs even better relative to other baselines. For instance, as shown in Table \ref{table:glue_debertav3-base}, at a target sparsity level of 90\%, MGPP achieves 85.2/84.2\% accuracy on the MNLI dataset—3.5/2.4\% higher than the best-performing baseline, LoSparse. Remarkably, our results at 90\% sparsity for MNLI even surpass LoSparse's performance at 80\% sparsity, demonstrating the effectiveness of our approach with more data and higher sparsity levels. Similarly, Table \ref{table:glue_bert-base} shows that our method, while using less memory, achieves better or comparable results to PLATON.

\begin{table*}[htbp]
\caption{Comparison of different pruning methods for the DeBERTaV3\textsubscript{base} model on the GLUE development sets, where 
``N.A.'' indicates non-convergence of the model, 
``m/mm'' denotes the accuracy for the matched and mismatched development sets of the MNLI task, and other metrics 
(i.e., Acc, F1, Mcc, P/S Corr) are defined in Table \ref{table:glue_statistics}. The highest-performing results for each dataset are highlighted in bold.} 
\label{table:glue_debertav3-base}
\begin{center}
\sisetup{detect-weight,mode=text}
\renewrobustcmd{\bfseries}{\fontseries{b}\selectfont}
\renewrobustcmd{\boldmath}{}
\begin{adjustbox}{max width=\textwidth}
\begin{tabular}{clcccccccc}
\toprule
\multirow{2}{*}{\textbf{Sparsity}} & \multirow{2}{*}{\textbf{Method}} & \textbf{MNLI} & \textbf{RTE} & \textbf{QNLI} & \textbf{MRPC} & \textbf{QQP} & \textbf{SST-2} & \textbf{CoLA} & \textbf{STS-B} \\ 
& & m/mm & Acc & Acc & Acc/F1 & Acc/F1 & Acc & Mcc & P/S Corr\\
\midrule
\bf0\% & DeBERTaV3\textsubscript{base} & 90.5/90.6 & 82.0 & 94.0 & 89.5/93.3 & 92.4/89.8 & 95.3 & 69.2 & 91.6/91.1\\
\midrule
\multirow{4}{*}{\bf80\%} 
    & MvP & N.A. & 61.2 & 86.0 & 79.2/85.0 & N.A. & 89.4 & N.A. & 84.3/84.3\\
    & ITP & 82.8/82.5 & N.A. & 87.8 & 82.0/87.0 & 90.0/86.4 & 90.8 & 49.0 & 87.4/87.0\\
    & LoSparse & 84.5/83.8 & 68.0 & 88.6 & 85.0/89.4 & 90.6/87.2 & 91.7 & 50.0 & 88.8/88.5\\
    & MGPP & \bf87.2/86.9 & \bf70.0 & \bf91.7 & \textbf{85.5}/89.4 & \bf91.3/88.3 & \bf93.2 & \bf56.3 & \bf88.9/88.5\\
\midrule
\multirow{4}{*}{\bf85\%} 
    & MvP & N.A. & 59.0 & N.A. & 78.5/84.3 & N.A. & 89.0 & N.A. & 83.9/83.9\\
    & ITP & 81.7/81.3 & N.A. & 85.4 & 80.5/86.3 & 89.1/85.2 & 89.3 & 45.8 & 86.8/86.3\\
    & LoSparse & 83.3/82.9 & 66.9 & 87.6 & 83.6/88.0 & 90.3/87.0 & 90.4 & 46.8 & 87.7/87.3\\
    & MGPP & \bf 86.0/85.9 & \bf68.3  & \bf90.9 & \bf84.3/88.7 & \bf91.1/87.9 & \bf92.3 & \bf50.9 & \bf88.0/87.5\\
\midrule
\multirow{4}{*}{\bf90\%} 
    & MvP & N.A. & N.A. & N.A. & 77.0/83.4 & N.A. & 88.0 & N.A. & N.A.\\
    & ITP & 79.7/79.6 & N.A. & 82.3 & 78.5/84.3 & 88.3/84.4 & 88.3 & 38.0 & 86.3/86.0\\
    & LoSparse & 81.7/81.8 & 66.0 & 86.1 & 82.3/\textbf{87.4} & 89.5/86.0 & 89.2 & 40.0 & 87.2/\textbf{87.0}\\
    & MGPP & \bf85.2/84.2 & \bf66.2 & \bf88.8 & \textbf{82.6}/87.1  & \bf91.1/88.0 & \bf90.2 & \bf48.0 & 87.2/86.7\\
\bottomrule
\end{tabular}
\end{adjustbox}
\end{center}
\end{table*}

\begin{table*}[htbp]
\caption{Comparison of different methods for the BERT\textsubscript{base} model on the GLUE development sets in downstream tasks, where "m/mm" denotes the accuracy for the matched and mismatched development sets of the MNLI task.
Refer to Table \ref{table:glue_statistics} for the other metrics used in the table.
}
\label{table:glue_bert-base}
\begin{center}
\sisetup{detect-weight,mode=text}
\renewrobustcmd{\bfseries}{\fontseries{b}\selectfont}
\renewrobustcmd{\boldmath}{}
\begin{adjustbox}{max width=\textwidth}
\begin{tabular}{c  l  c c c c  }
\toprule
\multirow{2}{*}{\textbf{Sparsity}} & \multirow{2}{*}{\textbf{Method}} & \textbf{MNLI} & \textbf{QQP} & \textbf{QNLI} & \textbf{SST-2} \\
& & m/mm & Acc/F1 & Acc & Acc\\
\midrule
\bf0\% & BERT\textsubscript{base} & 84.6 / 83.4 &  91.5 / 88.5  & 91.3 & 92.7 \\
\midrule
\multirow{4}{*}{\bf80\%}
& GMP      & 81.5 / 82.9 & 86.0 / 83.8 & 89.2 & 84.3 \\
& MvP & 81.6 / 82.1 & 90.6 / 87.5 & 88.3 & 89.0 \\
& PLATON   & 83.1 / 83.4 & 90.7 / 87.5 & 90.1 & 91.5 \\
& MGPP      & 83.1 / 83.4 & \textbf{90.8} / \textbf{87.6} & \textbf{90.2} & \textbf{91.9} \\
\midrule
\multirow{4}{*}{\bf90\%}
& GMP      & 78.8 / 79.0 & 78.8 / 77.0 & 86.6 & 80.7 \\
& MvP      & 80.7 / 81.1 & 90.2 / 86.7 & 86.6 & 87.4 \\
& PLATON   & 82.0 / 82.2 & 90.2 / 86.8 & 88.9 & 90.5 \\
& MGPP     & \textbf{82.1} / 82.2 & \textbf{90.4} / \textbf{87.1} & \textbf{89.2} & \textbf{90.8}\\
\bottomrule
\end{tabular}
\end{adjustbox}
\end{center}
\end{table*}

\subsection{Question Answering}

We assess the performance of MGPP on the DeBERTaV3\textsubscript{base} model \citep{he2021debertav3} by conducting experiments on a standard question answering dataset SQuADv1.1 \citep{rajpurkar2016squad}. 
SQuADv1.1 is a reading comprehension benchmark consisting of questions derived from Wikipedia articles, 
with 88k training samples and 10k validation samples.

For all sparsity levels, the number of training epochs and batch sizes is set to 10 and 16, respectively. We set the learning rate to \(5 \times 10^{-5}\), and for the MGP, we specify \( \sigma_0^2 = 1 \times 10^{-10} \) and \( \sigma_1^2 = 0.05 \). More details are given in the Appendix \ref{appendix:qs}.

The results on the SQuADv1.1 validation set are summarized in Table \ref{table:SQuAD-v1.1_debertav3-base} using two performance metrics: exact match (EM) and F1. All baseline results are taken directly from \citep{li2023losparse}.
 MGPP demonstrates performance that is either superior to or on par with existing methods across all sparsity levels.
Consistent with our findings on the GLUE benchmark, our method is especially effective in high sparsity regimes. For example, at the 90\% sparsity level, MGPP outperforms LoSparse (the best-performing baseline) by 5.1\% in terms of EM.

\begin{table*}[htbp]
\caption{Comparison of MGPP, ITP, and LoSparse for the DeBERTaV3\textsubscript{base} model on the  SQuADv1.1 validation set, where the best results for each dataset are highlighted in bold.} 
\label{table:SQuAD-v1.1_debertav3-base}
\begin{center}
\sisetup{detect-weight,mode=text}
\renewrobustcmd{\bfseries}{\fontseries{b}\selectfont}
\renewrobustcmd{\boldmath}{}
\begin{adjustbox}{max width=\textwidth}
\begin{tabular}{lc c c c c c }
\toprule
\multirow{2}{*}{\textbf{Dataset}} & \multicolumn{6}{c}{\textbf{SQuADv1.1}} \\ 
 & \multicolumn{6}{c}{EM/F1}\\
\midrule
\textbf{Sparsity} & \bf95\% & \bf90\% & \bf80\% & \bf70\% & \bf60\% & \bf50\% \\
\midrule
DeBERTaV3\textsubscript{base} &  \multicolumn{6}{c}{87.7/93.5}\\
\midrule
- ITP & 65.2/76.1 & 70.9/80.3 & 75.0/83.9 & 78.2/86.2 & 78.2/86.2 & 81.5/89.6 \\
- LoSparse & 69.3/79.1 & 72.9/82.8 & 76.8/85.8 & 80.2/88.0 & 82.1/89.4 & 82.3/90.3 \\
- MGPP & \bf73.7/82.9 & \bf78.0/86.2 & \bf80.2/88.6 & \bf81.1/89.5 & 82.1/\textbf{90.1} & \textbf{82.5}/90.3\\
\bottomrule
\end{tabular}
\end{adjustbox}
\end{center}
\end{table*}

\subsection{Natural Language Generation}

We assess the pruning performance of MGPP on the BART\textsubscript{large} model \citep{lewis2019bart} by conducting experiments on two natural language generation datasets: XSum \citep{narayan2018don} and CNN/DailyMail \citep{hermann2015teaching}. The objective is to generate either a concise summary or a highlight that captures the main point of a document.  
Refer to the Appendix \ref{appendix:nlg} for more detailed information about the datasets.

For all sparsity levels and both datasets, we set the number of training epochs to 12 and the batch size to 32.  The beam search length is fixed at 8, and the learning rate is set to \(2 \times 10^{-5}\). For the MGP, we set \( \sigma_0^2 = 1 \times 10^{-10} \) and \( \sigma_1^2 = 0.1 \). Additional details can be founded in the Appendix \ref{appendix:nlg}.

The results on the test sets of both datasets are summarized in Table \ref{table:BART-large_XSum_CNN/DailyMail}, using three performance metrics: ROUGE 1/2/Lsum scores \citep{lin2004rouge}. All baseline results are directly adopted from \citep{li2023losparse}. The comparison indicates that MGPP outperforms existing approaches across all sparsity levels on both datasets. Notably, the larger the performance gap between the fully fine-tuned dense model and its sparsified counterpart, the greater the extent to which MGPP outperforms the baselines. 
This trend is especially pronounced for the XSum dataset, where the higher task complexity leads to a more significant gap.

\begin{table}[htbp]
\caption{Comparison of MGPP, ITP, and LoSparse for the BART\textsubscript{large} model on the datasets: XSum and CNN/DailyMail, where  ``Lead-3''  represents choosing the first 3 sentences as the summary, and the best results for each dataset are highlighted in bold.} 
\label{table:BART-large_XSum_CNN/DailyMail}
\begin{center}
\sisetup{detect-weight,mode=text}
\renewrobustcmd{\bfseries}{\fontseries{b}\selectfont}
\renewrobustcmd{\boldmath}{}
\begin{adjustbox}{max width=\columnwidth}
\begin{tabular}{clcc}
\toprule
\textbf{Sparsity} & \textbf{Method} & \textbf{XSum} & \textbf{CNN/DailyMail} \\
\midrule
\multirow{2}{*}{\bf0\%} 
    & Lead-3 & 16.30/1.60/11.95 & 40.42/17.62/36.67 \\
    & BART\textsubscript{large} & 45.14/22.27/37.25 & 44.16/21.28/40.90\\
\midrule
\multirow{3}{*}{\bf50\%} 
    & ITP & 38.42/16.32/31.43 & 40.76/18.30/37.65 \\
    & LoSparse & 39.18/16.91/31.62 & 41.54/19.04/38.58\\
    & MGPP & \bf 42.92/19.70/34.80 & \bf 42.59/19.90/39.57\\
\midrule
\multirow{3}{*}{\bf60\%} 
    & ITP & 36.71/14.96/29.86 & 40.52/18.10/37.31 \\
    & LoSparse & 38.30/16.02/30.72 & 41.42/19.00/38.47\\
    & MGPP & \bf 41.69/18.75/33.69 & \bf 42.27/19.63/39.26\\
\midrule
\multirow{3}{*}{\bf70\%} 
    & ITP & 34.42/13.15/27.99 & 40.35/17.98/37.15 \\
    & LoSparse & 37.41/15.42/30.02 & 41.21/18.84/38.21\\
    & MGPP & \bf 40.20/17.33/32.34 & \bf 41.93/19.21/38.88\\
\bottomrule
\end{tabular}
\end{adjustbox}
\end{center}
\end{table}

\subsection{Upstream Pruning}
\label{section:upstream}


Upstream pruning \citep{zafrir2021prune} provides an alternative to the conventional downstream pruning approach. In upstream pruning, the model is pruned during the semi-supervised pre-training phase and then fine-tuned sparsely on specific downstream tasks. Models pruned in this manner generally exhibit improved generalization capabilities \citep{chen2020lottery, zafrir2021prune} 
and require significantly fewer computational resources for fine-tuning, 
as only the remaining parameters need to be adjusted. 
However, upstream pruning typically demands a considerably larger dataset compared to downstream pruning. Currently, oBERT \citep{kurtic2022optimal} stands as the state-of-the-art method for upstream pruning on BERT-like models and serves as the primary baseline for comparison in this section.

Following the guidelines established by oBERT, we use the BERT\textsubscript{base} model fine-tuned on two upstream datasets: BookCorpus and English Wikipedia. We then apply {MGPP} to prune the model on the same datasets for 3 epochs. Finally, we sparse-fine-tune the pruned model on the GLUE benchmark for 8 epochs. Detailed hyperparameters are provided in Appendix \ref{appendix:upstream_pruning}.

The results are presented in Table \ref{table:upstream_pruning}. We adopt all baseline results directly from \citet{kurtic2022optimal}. Notably, MGPP outperforms oBERT, despite the latter's additional memory requirement of ${O(50d)}$ and its greater computational complexity of ${O(mBd)}$.

\begin{table*}[htbp]
\caption{Comparison of MGPP and oBERT on development sets for the upstream-pruned model BERT\(_{\text{BASE}}\) at the 90\% sparsity level,
where "m/mm" indicates the accuracy for the matched and mismatched development sets of the MNLI task.}
\begin{center}
\sisetup{detect-weight,mode=text}
\renewrobustcmd{\bfseries}{\fontseries{b}\selectfont}
\renewrobustcmd{\boldmath}{}
\begin{adjustbox}{max width=\textwidth}
\begin{tabular}{c  l  c c c c c}
\toprule
\multirow{2}{*}{\textbf{Sparsity}} & \multirow{2}{*}{\textbf{Method}} & \textbf{MNLI} & \textbf{QNLI} & \textbf{QQP} & \textbf{SST-2} \\ 
& & m/mm & Acc &  Acc/F1 & Acc \\
\midrule
\bf0\% & BERT\textsubscript{BASE} & 84.6 / 83.4 & 91.3 & 91.5 / 88.5 & 92.7 \\
\midrule
\multirow{2}{*}{\bf90\%} 
    & oBERT & 82.2 / 82.5 & 89.3 & 90.4 / 87.1 & 92.0 \\
    & MGPP & \textbf{82.4} / \textbf{82.6} & \textbf{89.8} & \textbf{90.5} / \textbf{87.3} & \textbf{92.3} \\
\bottomrule
\end{tabular}
\end{adjustbox}
\end{center}
\label{table:upstream_pruning}
\end{table*}

\subsection{Ablation Study}
\label{main:ablation_study}

To justify the contributions of various components and design choices in our method, we conduct an ablation study in this section. We compare our approach to Prior-Annealing (PA) \citep{sun2021sparse, zhang2023sparse},
which provides an effective implementation for 
 sparsifying deep neural network models with the MGP prior
 in the one-shot pruning style.
Additionally, we replace the MGP in our method with an \( L_2 \) penalty (denoted as \(L_2\)) to confirm the significance of this particular prior. As we have previously discussed, the MGP imposes penalties on parameters in a way that is similar to \( L_2 \) regularization on a larger scale of the parameter space.

The ablation study is carried out on the DeBERTaV3\textsubscript{base} model on  three datasets from the GLUE benchmark: MNLI, MRPC, and SST-2. These datasets represent diverse task categories, including single-sentence classification, similarity and paraphrasing, and natural language inference. They also vary in training set size, ranked from large to small: MNLI, SST-2, MRPC.

The results are summarized in Table \ref{table:ablation}. Notably, we performed extensive hyperparameter search for PA to  ensure a fair comparison (details are given in Appendix \ref{appendix:as}). Despite this effort, MGPP consistently  outperforms PA on all three datasets and across all sparsity levels. When compared to \( L_2 \), 
the advantage of MGPP becomes increasingly pronounced as sparsity increases.
  For example, on the MNLI dataset, at 80\% sparsity, MGPP surpasses \( L_2 \) by 1.2/1.3\%. At 90\% sparsity, this margin grows significantly, with MGPP outperforming \( L_2 \) by 3.6/3.0\%.

\begin{table}[htbp]
\caption{Comparison of MGPP with two  ablation variants, PA and \( L_2 \), on the MNLI, MRPC, and SST-2 datasets, where the results of MGPP are taken from Table \ref{table:glue_debertav3-base}.} 
\label{table:ablation}
\begin{center}
\sisetup{detect-weight,mode=text}
\renewrobustcmd{\bfseries}{\fontseries{b}\selectfont}
\renewrobustcmd{\boldmath}{}
\begin{adjustbox}{max width=\columnwidth}
\begin{tabular}{clc c c }
\toprule
\multirow{2}{*}{\textbf{Sparsity}} & \multirow{2}{*}{\textbf{Method}} & \textbf{MNLI} & \textbf{MRPC} & \textbf{SST-2} \\ 
& & m/mm & Acc/F1 & Acc \\
\midrule
\multirow{3}{*}{\bf80\%} 
    & PA & 83.8/82.9 & 83.1/88.3 & 90.1 \\
    & $L_2$ & 86.0/85.6 & 82.4/87.3 & 91.5 \\
    & MGPP & \bf87.2/86.9 & \textbf{85.5/89.4} & \bf93.2 \\
\midrule
\multirow{3}{*}{\bf85\%}
    & PA & 81.6/81.4 & 78.4/85.7 & 88.5 \\
    & $L_2$ & 83.8/84.6 & 76.5/82.0  & 90.5 \\
    & MGPP & \bf 86.0/85.9 & \bf 84.3/88.7 & \bf92.3 \\
\midrule
\multirow{3}{*}{\bf90\%}
    & PA & 79.5/78.9 & 77.6/83.8 & 87.2 \\
    & $L_2$ & 81.6/81.2 & 71.8/82.1 & 87.1 \\
    & MGPP & \bf85.2/84.2 & \textbf{82.6/87.1}  & \bf90.2 \\
\bottomrule
\end{tabular}
\end{adjustbox}
\end{center}
\end{table}

\subsection{Algorithm Analysis}

\begin{figure}[!ht]
  \centering
  \begin{tabular}{cc}
  (a) & (b) \\ 
  \includegraphics[width=0.45\linewidth]{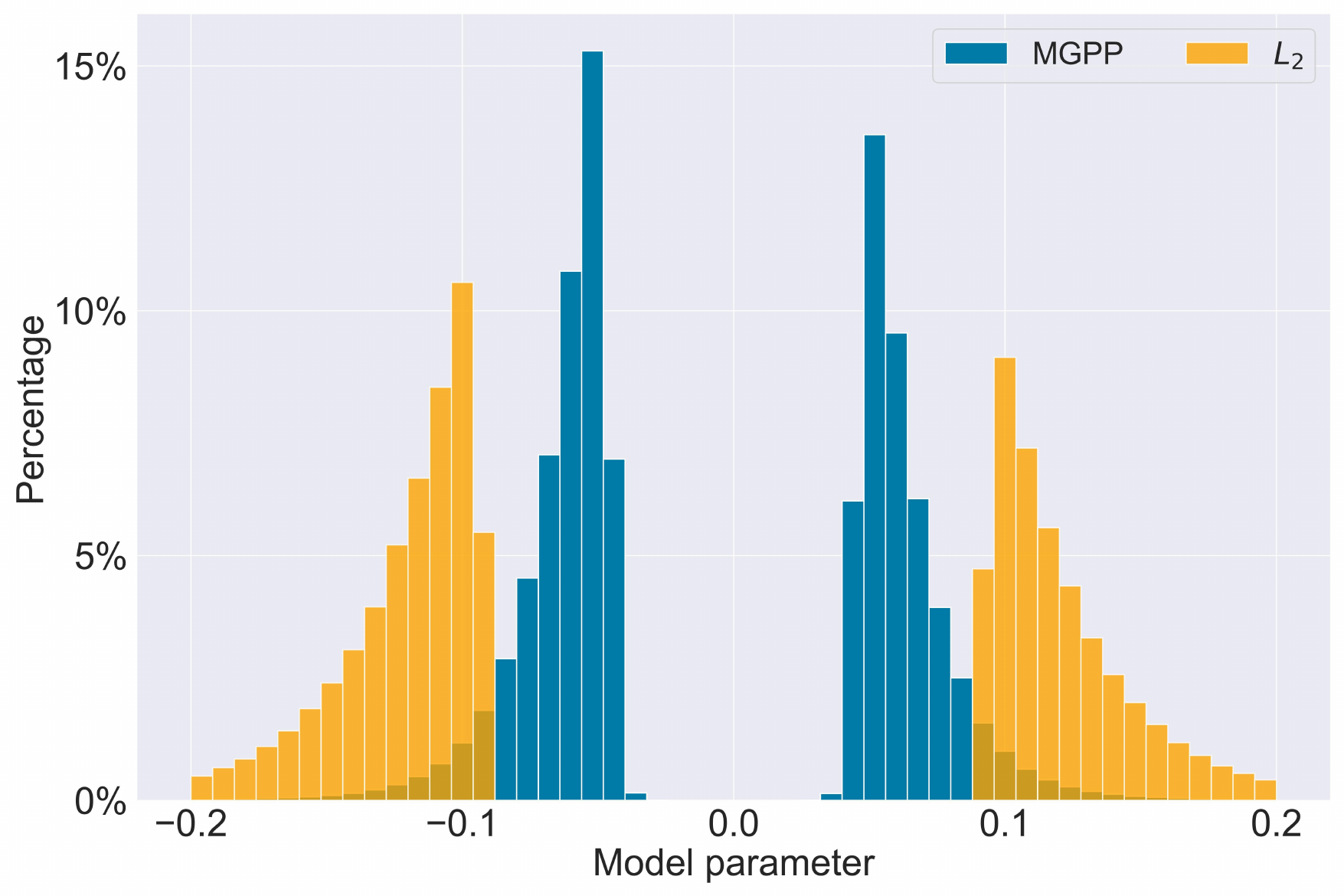} & 
  \includegraphics[width=0.45\linewidth]{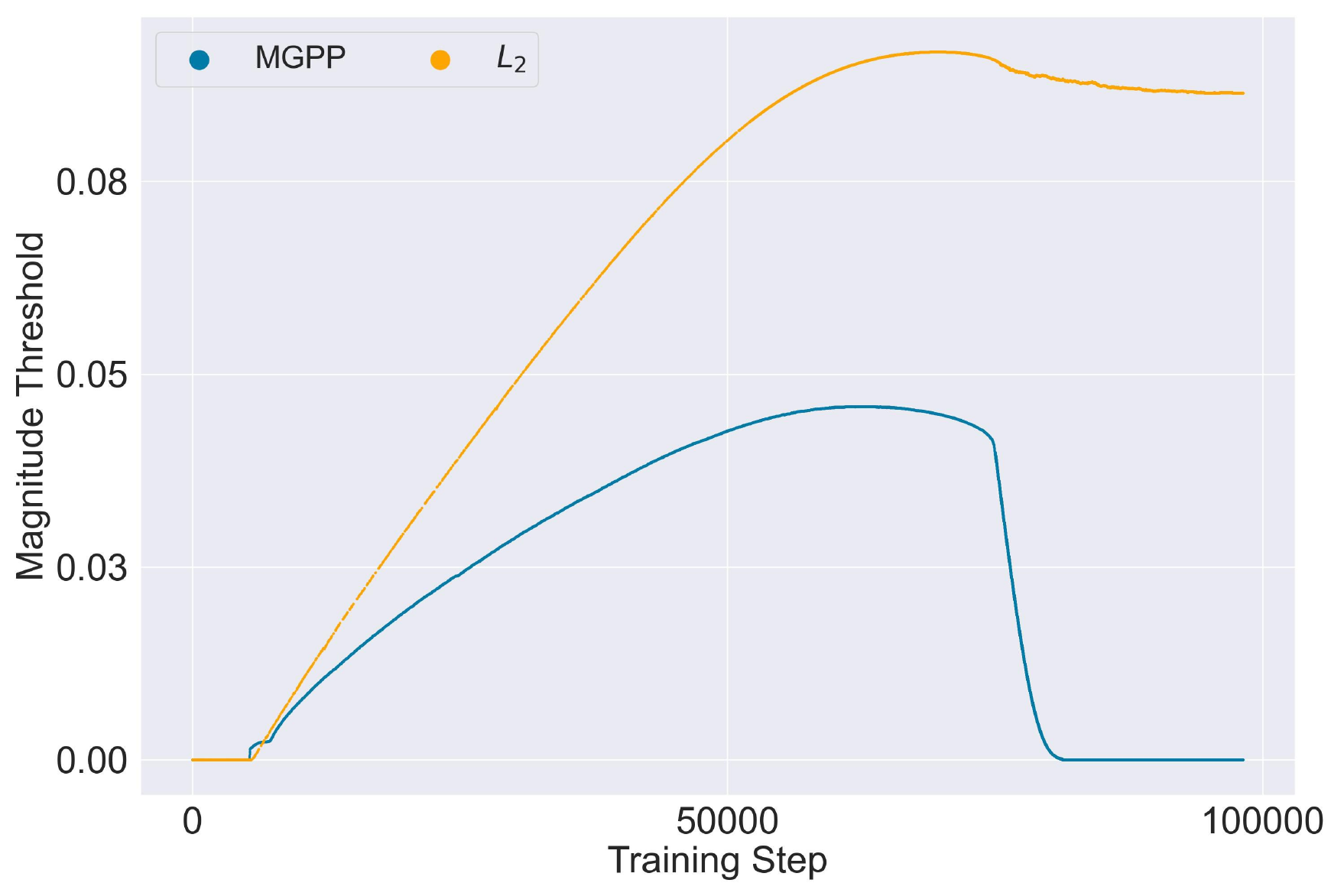}
  \end{tabular} 
  \caption{A comparative analysis of MGPP and \( L_2 \): (a)   distributions of remaining nonzero parameters,  (b) magnitude pruning thresholds during training.}
  \label{fig:mgpp_vs_l2}
\end{figure}

To better illustrate the impact of MGP, Figure \ref{fig:mgpp_vs_l2} depicts the distribution of remaining nonzero parameters and the evolution of pruning thresholds during training for a 90\% sparsified DeBERTaV3\textsubscript{base} model on the MNLI dataset, comparing our method against the \( L_2 \) variant. We observe that both MGPP and \( L_2 \) tend to prune parameters that are close to zero. 
However, as shown in Figure \ref{fig:mgpp_vs_l2}(b), the spike component in the MGP more effectively 
drives parameters toward zero, resulting in  a lower pruning threshold. In contrast, \( L_2 \) fails to similarly  reduce the pruning threshold, leading to  a performance gap in generalization.

\subsection{Hyperparameter Sensitivity Analysis}
\label{hs}


The proposed method, MGPP, introduces six hyperparameters: three from the Mixture Gaussian Prior (MGP) and three from the cubic sparsity scheduler. It is important to note that the cubic sparsity scheduler is also employed by the baselines considered in this work, so no additional hyperparameters are introduced when comparing to these baselines.

In this section, we focus on the sensitivity of the three MGP-specific hyperparameters:

\begin{itemize}
    \item \textbf{$\lambda$}: MGPP is robust to this hyperparameter, which we fix at $1\times10^{-7}$ in all experiments. Changing $\lambda$ only slightly adjusts the width of the spike component (see Appendix \ref{appendix:mgps}). Preliminary experiments show that its impact on performance is negligible, and a value below 0.1 is generally sufficient.

    \item \textbf{$\sigma_0^2$}: A general guideline is to use a smaller value when more training samples are available. We limited our selection to the set $\{1\times10^{-9}, 1\times10^{-10}\}$. Values in the range $1\times10^{-12} \leq \sigma_0^2 \leq 1\times10^{-8}$ do not significantly affect performance.

    \item \textbf{$\sigma_1^2$}: Similar to $\sigma_0^2$, smaller values are recommended for larger datasets. We restricted our choices to the set $\{0.1, 0.05\}$. Although this hyperparameter has more impact on performance, the suggested values work well across all experiments.
\end{itemize}

For more detailed discussion on how these hyperparameters shape the prior landscape, please refer to Appendix \ref{appendix:mgps}.



\section{Conclusion}

In this paper, we have developed MGPP, a novel magnitude-based iterative pruning method designed 
 to sparsify large-scale transformer models. 
Extensive experimental results on various natural language processing tasks and two transformer-based language models demonstrate the effectiveness and efficiency of  MGPP,
 particularly in settings with abundant   training data  or  high sparsity. Additionally, we provided a theoretical justification for the consistency of MGPP, offering insights into its strong performance.

Transformer model pruning is an ongoing research area. Besides the pruning scores discussed in Section \ref{main:pruning_scores}, more complex pruning scores have also been proposed in the literature.
For instance, 
the Platon method \citep{zhang2022platon} prunes the model based on the upper confidence bound of the weight importance, while  
the WoodFisher \citep{singh2020woodfisher}, M-FAC \citep{frantar2021m},
and oBERT \citep{kurtic2022optimal} methods utilize Hessian-based information for pruning. 
These pruning scores can also be computed with the mixture Gaussian prior, leading to new variants of the proposed method. Notably, 
the consistency property of the MGPP method can be extended to these new variants, providing a theoretical guarantee for their validity. In contrast, existing methods often lack 
such theoretical support for their performance.
Additionally, we note that the calculation of complex pruning scores often requires higher GPU memory than that needed for magnitude-based pruning scores.

While model compression often involves other strategies like knowledge distillation and quantization, these are not mutually exclusive with pruning. For instance, one could enhance the performance of a pruned model through knowledge distillation and further reduce storage requirements by quantizing the remaining parameters. We leave such extensions for future work.

\section*{Acknowledgement}

 The authors thank the editor, associate editor, and referees for their constructive comments which has led to significant improvement of this paper. 
 Liang's research is support in part by the NSF grants DMS-2015498 and DMS-2210819, and the NIH grant R01-GM152717.

\vspace{0.5in}
 
 \appendix

\begin{center}
{\bf \Large Appendix}
\end{center}

\setcounter{section}{0}
\renewcommand{\thesection}{A\arabic{section}}
\setcounter{table}{0}
\renewcommand{\thetable}{A\arabic{table}}
\setcounter{figure}{0}
\renewcommand{\thefigure}{A\arabic{figure}}
\setcounter{equation}{0}
\renewcommand{\theequation}{A\arabic{equation}}
\setcounter{lemma}{0}
\renewcommand{\thelemma}{A\arabic{lemma}}
\setcounter{theorem}{0}
\renewcommand{\thetheorem}{A\arabic{theorem}}
\setcounter{remark}{0}
\renewcommand{\theremark}{A\arabic{remark}}


The appendix is organized as follows. Section \ref{Transformersection}  provides a loose justification for the consistency of the proposed MGPP method. In Section \ref{auxsection}, we introduce an auxiliary stochastic transformer model; and in Section \ref{IROsparsedecodersection}, we  provide a constructive proof for the consistency of the sparse stochastic transformer estimator under the theoretical framework of the imputation regularized-optimization (IRO) algorithm \citep{Liang2018missing}. 
With this preparation, 
we then establish the consistency of the sparse transformer model with a mixture Gaussian prior/penalty. 
Section \ref{appendix:mgps} provides visualization of how \( \lambda \), \( \sigma_0^2 \), and \( \sigma_1^2 \) affect the landscape of the MGP.
 Section \ref{appendix:pa} gives details of the prior annealing
 Section \ref{appendix:hs} provides sensitivity analysis for hyperparameters. 
 Section \ref{appendix:all_exp} provides settings for the experiments.

\section{Consistency of Sparse Transformer with the MGP Penalty} \label{Transformersection}

\subsection{Asymptotic Equivalence between the Transformer and Stochastic Transformer Models} \label{auxsection}

The asymptotic equivalence between 
the transformer and stochastic transformer 
has been studied in \cite{KimSLiang2024Narrow}. To make the paper 
self-contained, we provided a brief review as follows.

\paragraph{Transformer Model} 
Following \cite{Thickstun2020TheTM}, we define  a transformer block as follows.  
Let $\bx=(\bx_1, \bx_2,\ldots,\bx_n)^T \in \mathbb{R}^{n\times d}$ denote an input matrix to a transformer block, which transforms 
$\bx$ to  $\bz\in \mathbb{R}^{n\times d}$ 
with the detail specified as follows: 
 \begin{equation} \label{transformereq}
 \begin{split} 
 & Q^{(h)}(\bx_i)=W_{h,q}^T\bx_i, \ \ K^{(h)}(\bx_i)=W_{h,k}^T \bx_i,  \\
 &  V^{(h)}(\bx_i)=W_{h,v}^T \bx_i, \ \ W_{h,q}, W_{h,k}, W_{h,v} \in \mathbb{R}^{d\times k}, \\ 
& {
\alpha_{i,j}^{(h)}={\rm softmax}_j \left( \frac{ \langle Q^{(h)}(\bx_i),  K^{(h)}(\bx_j)\rangle}{\sqrt{k}} \right)},  \ i,j=1,2,\ldots,n, \\ 
& \bu_i=\sum_{h=1}^H W_{c,h}^T \sum_{j=1}^n \alpha_{i,j}^{(h)} V^{(h)}(\bx_j), \quad 
W_{c,h} \in \mathbb{R}^{k\times d}, \\ 
&\tilde{\bu}_i={\rm LayerNorm}(\bx_i+\bu_i; \bgamma_1,\bbeta_1), \quad \bgamma_1,\bbeta_1 \in \mathbb{R}^d, \\ 
& \tilde{\bz}_i=W_2^T {\rm ReLU}(W_1^T \tilde{\bu}_i), \quad W_1\in \mathbb{R}^{d\times m}, \ W_2 \in \mathbb{R}^{m\times d}, \\ 
& \bz_i={\rm LayerNorm}(\tilde{\bu}_i+\tilde{\bz}_i; \bgamma_2,\bbeta_2), \quad \bgamma_2,\bbeta_2 \in \mathbb{R}^d, \\
\end{split}
\end{equation} 
 where $H$ denotes the number of attention heads, $\langle\cdot,\cdot \rangle$ denotes inner product, and the layerNorm is given by 
 \[
 {\rm LayerNorm}(\ba; \bgamma,\bbeta)= \bgamma \odot \frac{(\ba-\bar{a}{\bf 1}_d)}{s_{a}}+\bbeta, 
 \]
 where $\odot$ is the element-wise multiplication operator, 
 $\ba=(a_1,a_2,\ldots, a_d)^T \in \mathbb{R}^d$,   
 ${\bf 1}_d$ is an $d$-vector of ones,  $\bar{a}=\frac{1}{d} \sum_{i=1}^d a_i$, 
 and $s_{a}=\sqrt{\frac{1}{d} \sum_{i=1}^d (a_i-\bar{a})^2}$. 
 For convenience, we denote the transformer block by the function $f_{\btheta}: \mathbb{R}^{n\times d} \to \mathbb{R}^{n\times d}$, where  
 the parameters $\btheta$ consist of  $\{W_{h,q}, W_{h,k}, W_{h,v}, W_{c,h}: h=1,2,\ldots,H\}$
  and $\{\bgamma_1,\bbeta_1,\bgamma_2,\bbeta_2, W_1, W_2\}$. 
  A transformer is a composition of $L$ transformer blocks:  
  $f_{\btheta_L} \circ \cdots \circ f_{\btheta_1}(\bx)\in \mathbb{R}^{n\times d}$, each block
  has its own parameters.  The common settings of the hyperparameters are 
  $d=512$, $k=64$, $m=2048$, and $H=8$.

 \paragraph{Stochastic Transformer Model} 
  The stochastic transformer model \citep{KimSLiang2024Narrow} is defined as follows: 
 \begin{equation} \label{transformereq:stochastic}
 \begin{split} 
 & Q^{(h)}(\bx_i)=W_{h,q}^T \bx_i+\bepsilon^{h,q}, \ \ K^{(h)}(\bx_i)=W_{h,k}^T \bx_i+\bepsilon^{h,k},  \\
 &  V^{(h)}(\bx_i)=W_{h,v}^T \bx_i+\bepsilon^{h,v}, \ \ W_{h,q}, W_{h,k}, W_{h,v} \in \mathbb{R}^{d\times k}, \\ 
& {
\alpha_{i,j}^{(h)}={\rm softmax}_j \left( \frac{ \langle Q^{(h)}(\bx_i),  K^{(h)}(\bx_j)\rangle}{\sqrt{k}} \right)},  \ i,j=1,2,\ldots,n, \\ 
& \bu_i=\sum_{h=1}^H W_{c,h}^T \sum_{j=1}^n \alpha_{i,j}^{(h)} V^{(h)}(\bx_j) +\bepsilon^{i,u}, \quad W_{c,h} \in \mathbb{R}^{k\times d}, \\ 
&\tilde{\bu}_i={\rm LayerNorm}(\bx_i+\bu_i; \bgamma_1,\bbeta_1), \quad \bgamma_1,\bbeta_1 \in \mathbb{R}^d, \\ 
& \tilde{\bz}_i=W_2^T {\rm ReLU}(W_1^T \tilde{\bu}_i) +\bepsilon^{i,\tilde{z}}, \  W_1\in \mathbb{R}^{d\times m}, \ W_2 \in \mathbb{R}^{m\times d}, \\ 
& \bz_i={\rm LayerNorm}(\tilde{\bu}_i+\tilde{\bz}_i; \bgamma_2,\bbeta_2), \quad \bgamma_2, \bbeta_2 \in \mathbb{R}^d, \\
\end{split}
\end{equation} 
where the noise variables $\bepsilon^{h,q} \sim N(0, \sigma_{q}^2 I_k)$, $\bepsilon^{h,k} \sim N(0, \sigma_{k}^2 I_k)$, $\bepsilon^{h,v} \sim N(0, \sigma_{v}^2 I_k)$, 
$\bepsilon^{i,u} \sim N(0, \sigma_{u}^2 I_d)$, 
and $\epsilon^{i,\tilde{z}} \sim N(0, \sigma_{\tilde{z}}^2 I_d)$ 
are mutually independent. 
Note that $\sigma_{q}^2$, $\sigma_{k}^2$, $\sigma_{v}^2$, 
$\sigma_{u}^2$, and $\sigma_{\tilde{z}}^2$ are all known, pre-specified by user. 
As a consequence of introducing the noise variables, we can treat $Q^{(h)}$'s, $K^{(h)}$'s, 
 $V^{(h)}$'s, $\bu_i$'s, and $\tilde{\bz}_i$'s as latent variables, 
 and  decompose the model as 
\begin{equation} \label{decompeq}
\begin{split} 
 & \pi_{\btheta}(\bz, \bQ, \bK,\bV,\bU, \tilde{\bZ}|\bx)
 = \prod_{h=1}^H \pi(Q^{(h)}|\bx,\btheta^{(1)})   
  \prod_{h=1}^H \pi(K^{(h)}|\bx,\btheta^{(2)}) \prod_{h=1}^H \pi(V^{(h)}|\bx,\btheta^{(3)}) \\
 & \quad  \times \pi(\bU|\bx,\btheta^{(4)},\bQ,\bK,\bV) \pi(\tilde{\bZ}|\bx,\btheta^{(5)},\bQ,\bK,\bV,\bU) 
 \pi(\bz|  \bQ,\bK,\bV,\bU,\tilde{\bZ}, \bx,\btheta^{(6)}),  \\
 \end{split} 
\end{equation}
where $\bQ=\{Q^{(1)}, Q^{(2)}, \ldots, Q^{(H)}\}$, $\bK=\{K^{(1)}, K^{(2)}, \ldots, K^{(H)}\}$,
$\bV=\{V^{(1)}, V^{(2)}, \ldots, V^{(H)}\}$, $\bU=\{\bu_1,\bu_2,\ldots,\bu_n\}$, 
$\tilde{\bZ}=\{\tilde{\bz}_1,\tilde{\bz}_2,\ldots,\tilde{\bz}_n\}$,
$\btheta^{(1)}=\{W_{h,q}: h=1,2,\ldots,H\}$, $\btheta^{(2)}=\{W_{h,k}: h=1,2,\ldots,H\}$, 
$\btheta^{(3)}=\{W_{h,v}: h=1,2,\ldots,H\}$, $\btheta^{(4)}=\{W_{c,h}: h=1,2,\ldots,H\}$,  
$\btheta^{(5)}=\{W_1, W_2, \bgamma_1,\bbeta_1\}$, 
and $\btheta^{(6)}=\{\bgamma_2,\bbeta_2\}$. 

The asymptotic equivalence between the transformer and stochastic transformer models have been established in \cite{KimSLiang2024Narrow}, where it was shown that  
the two models have asymptotically the same loss function under appropriate conditions. 
More precisely, they showed that
there exists  a small value $\tau(d,k,m,H)$, 
as a function of $d$, $k$, $m$ and $H$, such that 
\begin{equation} \label{equivLikelihoodeq1}
\begin{split}
\sup_{\btheta\in \Theta} & \frac{1}{n}\Big| \log \pi_{\btheta} (\bz, \bQ,\bK,\bV,\bU,\tilde{\bZ}|\bx) 
- \log \tilde{\pi}_{\btheta} (\bz|\bx) \Big| \stackrel{p}{\to} 0, 
\end{split}
\end{equation}
as $n \to \infty$ and $\max\{\sigma_q, \sigma_k, \sigma_v,\sigma_u,\sigma_{\tilde{z}}\} \prec \tau(d,k,m,H)$, where  
$\pi_{\btheta} (\bz, \bQ,\bK,\bV,\bU,\tilde{\bZ}|\bx)$ represents the pseudo-complete data likelihood function of the stochastic transformer by treating 
$\bQ$, $\bK$, $\bV$, 
$\bU$, and $\tilde{\bZ}$ as latent variables, 
$\tilde{\pi}_{\btheta}(\bz|\bx)$ represents  the likelihood function 
of the  transformer,  and 
$\stackrel{p}{\to}$ denotes convergence in probability.
We note that similar techniques have been used in  \cite{LiangSLiang2022StoNet} and \cite{SunLiang2022kernel} in establishing the asymptotic equivalence between the deep neural network and stochastic neural network (StoNet) models.

\subsection{Consistency of Sparse Transformer} \label{IROsparsedecodersection}

 By treating $\{\bQ, \bK, \bV, \bU, \tilde{\bZ}\}$ as latent variables, 
 the parameters 
$\btheta$ of the stochastic transformer model can be estimated using 
a regularization approach as follows: 
 \begin{equation} \label{misseq1}
 \begin{split}
 \hat{\btheta}_n & =\arg\max_{\btheta}\Big\{ \log \pi_{\btheta} (\bz, \bQ, \bK, \bV,\bU, \tilde{\bZ}|\bx) +\log P_{\lambda} (\btheta) \Big\},
 \end{split}
 \end{equation} 
 where $P_{\lambda}(\btheta)$ denotes the sparsity penalty imposed on $\btheta$, and $\lambda$ is the tuning parameter. 
With an appropriate choice of $P_{\lambda}(\btheta)$, we can provide 
 a constructive proof for the consistency of $\hat{\btheta}_n$ 
based on the IRO algorithm \citep{Liang2018missing}. 

The IRO algorithm starts with an initial weight setting $\hat{\btheta}_n^{(0)}$ 
and then iterates between the imputation and regularized optimization steps: 
\begin{itemize}
    \item {\bf Imputation:}  For each block, conditioned on the current parameter estimate $\btheta_{t-1}$, simulate the latent variables  $(\bQ,\bK,\bV,\bU,\tilde{\bZ})$ 
     from the predictive  distribution 
\begin{equation} \label{decompeq2}
\begin{split} 
 & \pi_{\btheta_{t-1}}(\bQ_{t}, \bK_t, \bV_t,\bU_t,\tilde{\bZ}_t|\bx,\bz) \propto \prod_{h=1}^H \pi(Q_t^{(h)}|\bx,\btheta_{t-1}^{(1)}) 
\prod_{h=1}^H \pi(K_t^{(h)}|\bx,\btheta_{t-1}^{(1)})  
  \prod_{h=1}^H \pi(V_t^{(h)}|\bx,\btheta_{t-1}^{(2)})  
\\ &\times 
\pi(\bU_t|\bx,\btheta_{t-1}^{(4)},\bQ_t,\bK_t,\bV_t) 
\pi(\tilde{\bZ}_t|\bx,\btheta_{t-1}^{(5)},\bQ_t,\bK_t,\bV_t,\bU_t) 
\pi(\bz|  \bQ_t,\bK_t,\bV_t,\bU_t,\tilde{\bZ}_t, \bx,\btheta_{t-1}^{(6)}),  \\
 \end{split} 
\end{equation}
 where $t$ indexes iterations, $\bQ_t=\{Q_t^{(h)}: h=1,2,\ldots,H\}$,  $\bK_t=\{K_t^{(h)}: h=1,2,\ldots,H\}$, $\bV_t=\{ V_t^{(h)}: h=1,2,\ldots,H\}$,
$\bU_t=\{\bu_{1,t},\bu_{2,t},\ldots,\bu_{n,t}\}$, 
$\tilde{\bZ}_t=\{\tilde{\bz}_{1,t},\tilde{\bz}_{2,t},\ldots,\tilde{\bz}_{n,t}\}$.
Here, $\bu_{i,t}$ and $\tilde{\bz}_{i,t}$ denote, respectively, the imputed values for 
$\bu_i$  and $\bz_i$ at iteration $t$. 
 
  \item {\bf Regularized optimization:} Given the pseudo-complete data $\{ \bQ_{t}, \bK_t, \bV_t, \bU_t,\tilde{\bZ}_t, \bz, \bx\}$, update  $\hat{\btheta}_n^{(t-1)}$ by maximizing the penalized log-likelihood function as follows: 
  \begin{equation} \label{IROsolution}
  \begin{split}
  \hat{\btheta}_n^{(t)} & =\arg\max_{\btheta}\Big\{ \log \pi_{\btheta}(\bz, 
  \bQ_{t}, \bK_t, \bV_t, \bU_t,\tilde{\bZ}_t |\bx)   
  +\log P_{\lambda} (\btheta) \Big\}, 
 \end{split}
\end{equation}
 which, by the decomposition (\ref{decompeq}), can be reduced to solving 
 for  $\btheta^{(1)}, \ldots, \btheta^{(6)}$, separately. 
 The penalty function $P_{\lambda}(\btheta)$ should be chosen such that $\hat{\btheta}_n^{(t)}$ forms a consistent estimator for the working parameter 
 \[
 \begin{split}
 &\btheta_*^{(t)} =\arg\max_{\btheta} \mathbb{E}_{\hat{\btheta}^{(t-1)}} 
 \log \pi(\bZ,\bQ_t,\bK_t,\bV_t,\bU_t,\tilde{\bZ}_t|\btheta,\bx) \\
 & = \arg\max_{\btheta}   
 \int \log \pi(\bz, \bQ_t,\bK_t,\bV_t, \bU_t,\tilde{\bZ}_t|\btheta,\bx) \\ 
 &  \quad \times \pi(\bQ_t,\bK_t,\bV_t, \bU_t,\tilde{\bZ}_t|\bz,\bx,\btheta_n^{(t-1)}) \pi(\bz|\bx,\btheta^*) 
 d\bQ_t d \bK_t  d \bV_t d\bz, 
 \end{split}
 \]
 where   $\btheta^*$ is defined by  
 \begin{equation} \label{trueparameter}
 \btheta^*=\arg\max_{\btheta} \mathbb{E} \log \pi(\bz|\btheta,\bx),
 \end{equation}
 and it corresponds to  the true parameters of the  underlying sparse transformer model (\ref{transformereq}). 
 \end{itemize}

For the sake of theoretical simplicity, we can assume that the hyperparameters of the transformer, namely, $d$, $k$, $m$ and $H$, can increase with $n$ but at a low order. 
By standard statistical estimation theory, see e.g., \cite{Portnoy1988}, 
we can achieve consistency of $\hat{\btheta}_n^{(t)}$  
with the mixture Gaussian prior/penalty at each iteration of the IRO algorithm.  

The above assumption can be much relaxed. For example, we may assume that
$d$, $k$, $m$ and $H$ increase with $n$ exponentially. Under this extended assumption, we can still achieve consistency of $\hat{\btheta}_n^{(t)}$  
with the mixture Gaussian prior/penalty   
at each iteration of the IRO algorithm. This is possible by leveraging the theories presented in 
\cite{Song2017NearlyOB}, \cite{sun2021consistent}, and \cite{sun2021sparse}. 
To elaborate,  the estimation of $\btheta^{(1)}, \ldots,\btheta^{(4)}$ is reduced to solving a series of high-dimensional linear regressions, 
for which consistency with the mixture Gaussian prior 
can be maintained, as per the theory from \cite{Song2017NearlyOB}. 
The estimation of 
$\btheta^{(5)}$ is reduced to solving a sparse deep neural network model with 
the ReLU and linear activation functions, ensuring consistency with the mixture Gaussian prior  based on the theories outlined in \cite{sun2021consistent} and \cite{sun2021sparse}. 
The case of $\btheta^{(6)}$ is similar to  $\btheta^{(5)}$, it is reduced to solving a 
sparse deep neural network model when a fully connected neural network is added to connect 
$\bz$ and $\by$. Otherwise, the parameters $\{\bgamma_2,\bbeta_2\}$ can be uniquely determined. 
Note that, as mentioned in the main text, the parameters in the LayerNorm transformation are not sparsified. 

Furthermore, according to Theorem 4 of \cite{Liang2018missing}, 
the estimator $\hat{\btheta}_n^{(t)}$ is consistent when both $n$ and $t$ are sufficiently large.
In summary, under mild regularity conditions and the mixture Gaussian prior, 
we can establish that 
\begin{equation} \label{IROthetaconsistency1}
\|\hat{\btheta}_n^{(t)} - \btheta^* \| \stackrel{p}{\to} 0, 
\end{equation} 
for sufficiently large $n$ and sufficiently large $t$ and almost every dataset 
$\{\bx,\by\}$ for the stochastic transformer model. 

Finally, based on (\ref{equivLikelihoodeq1}) and under certain regularity 
conditions as given in 
\cite{LiangSLiang2022StoNet}, we also have 
\begin{equation} \label{IROthetaconsistency2}
\|\tilde{\btheta}_n^{(t)} - \btheta^* \| \stackrel{p}{\to} 0,
\end{equation} 
where $\tilde{\btheta}_n^{(t)}$ is a sparse transformer estimator give by 
\begin{equation} \label{sparseTransest}
\tilde{\btheta}_n^{(t)}=\arg\max_{\btheta}\Big\{ \log f_{\btheta}(\bz|\bx)+ \log P_{\lambda}(\btheta) \Big\}.
\end{equation}

In summary, through the introduction of an auxiliary stochastic transformer model and 
the utilization of the IRO convergence theory, we have justified the consistency of the sparse transformer model under mild regularity conditions similar to those given 
in \cite{LiangSLiang2022StoNet} and \cite{Liang2018missing}.

Finally, we note that the above justification for the consistency of the sparse transformer model is based on the assumption that $\bx \in \mathbb{R}^{n\times d}$ consists of $n$ i.i.d observations. In practice, the observations might exhibit correlations. Nevertheless, this should not significantly impact the validity of our results, as long as $\bx$ contains a sufficiently large number of independent samples.

\section{Mixture Gaussian Priors}
\label{appendix:mgps}

\begin{figure}[htpb]
  \centering
  \begin{tabular}{ccc} 
  (a)  & (b) & (c) \\ 
  \includegraphics[width=0.3\linewidth]{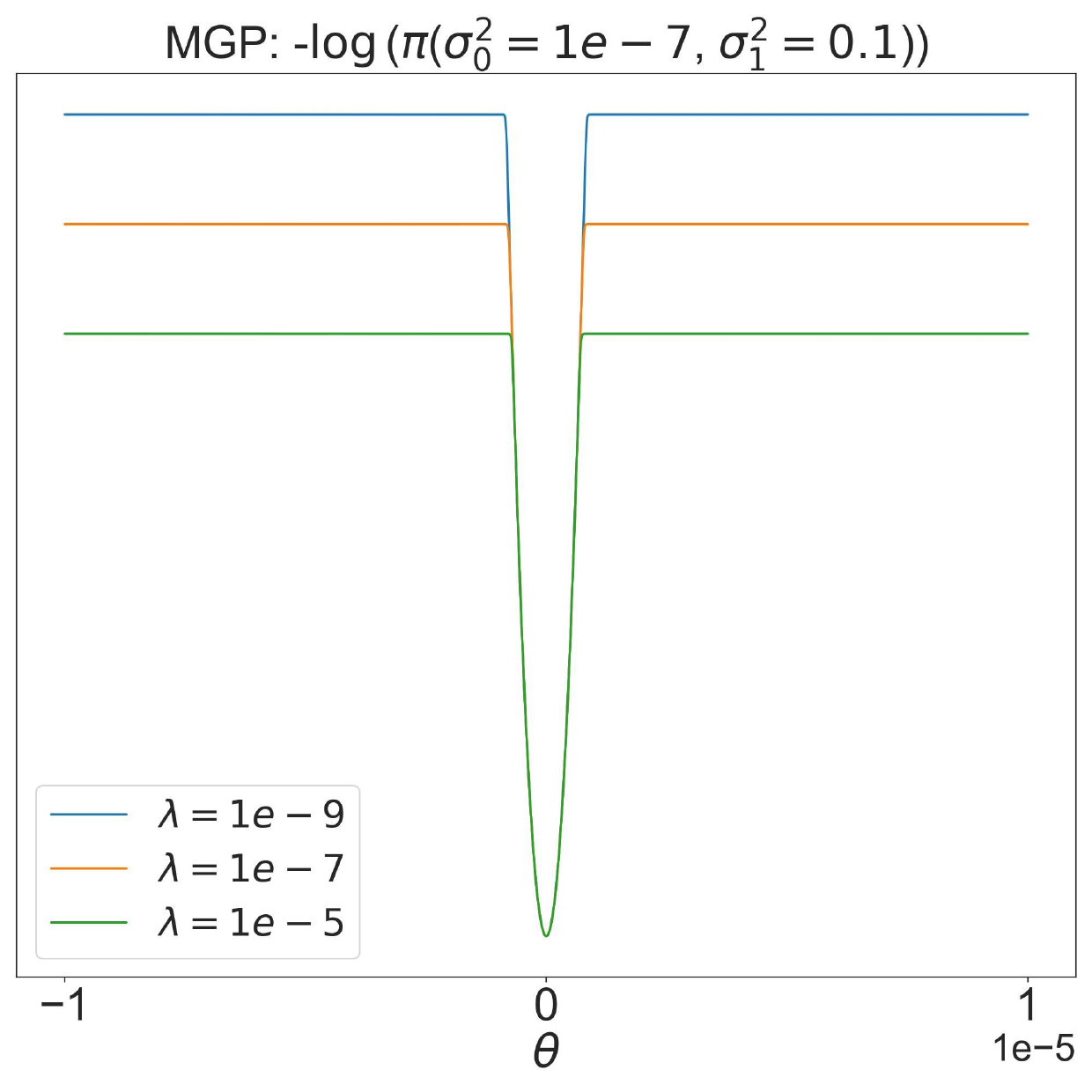} & 
    \includegraphics[width=0.3\linewidth]{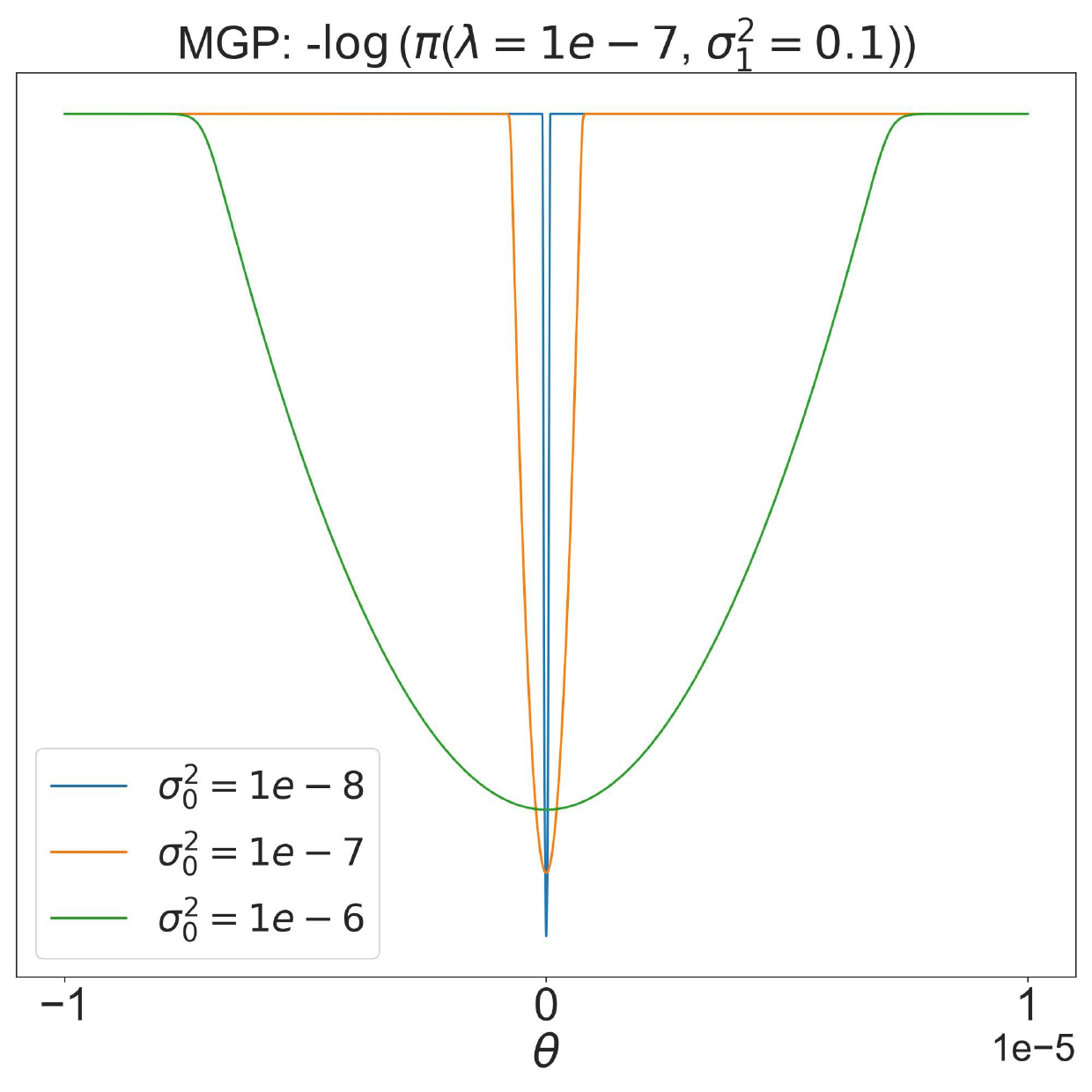} & 
    \includegraphics[width=0.3\linewidth]{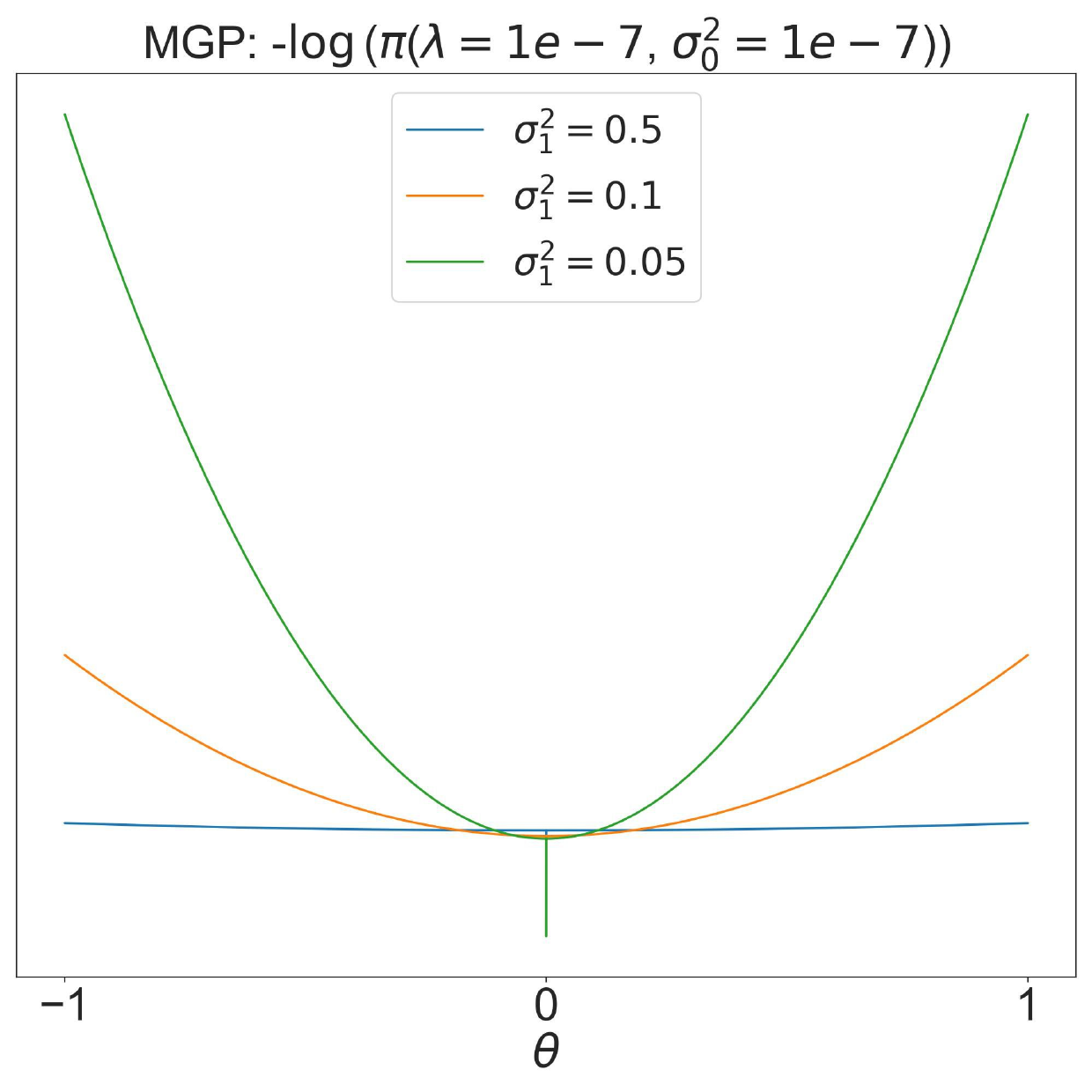}
 \end{tabular}
  \caption{How $\lambda$, $\sigma_0^2$, and $\sigma_1^2$ change the landscape of the MGP.}
  \label{fig:mgp_sigam0_sigma1}
\end{figure}

In this section, we illustrate and visualize how changes in $\lambda$, $\sigma_0^2$, and $\sigma_1^2$ influence the landscape of the MGP. As shown in Figure \ref{fig:mgp_sigam0_sigma1} (a), the effect of $\lambda$ primarily impacts the spike component of the MGP. The larger the value of $\lambda$, the wider the spike component becomes. Based on Figure \ref{fig:mgp_sigam0_sigma1} (b), the influence of $\sigma_0^2$ is most noticeable on the parameter space near zero. A smaller value of $\sigma_0^2$ results in a greater penalty applied to parameters within the spike component while making it smaller. Conversely, as depicted in Figure \ref{fig:mgp_sigam0_sigma1} (c), the impact of $\sigma_1^2$ is mainly observed in larger-scale areas. A smaller value of $\sigma_1^2$ imposes a higher penalty on parameters at a larger scale.


\section{Prior Annealing}
\label{appendix:pa}

In this section, we provide a brief overview of how the Prior-Annealing (PA) algorithm \citep{sun2021sparse, zhang2023sparse} is employed for model pruning. The detailed steps are outlined in Algorithm \ref{alg: pa}.

\begin{algorithm}[htbp]
   \caption{Prior-Annealing (PA)}
   \label{alg: pa}
\begin{algorithmic}[1]
    \State \textbf{Input:} training dataset $D_n$, pretrained model $\btheta^{(0)}$, number of training epochs $E$, mini-batch size $m$, $\lambda$, $\sigma_1^2$, $(\sigma_0^{\text{init}})^2$, $(\sigma_0^{\text{end}})^2$, initial temperature $\tau^{(0)}$, $t_i$, $t_f$.

    \State \textbf{Initialize:} $t=1$, $T = \ceil*{En/m}$, a stochastic gradient MCMC (SGMCMC) optimizer \citep{ma2015complete}.
    
    \State \textbf{Annealing the Prior:} Initialize $\btheta^{(t)}$ at $\btheta^{(0)}$, and simulate from a sequence of distributions $\pi(\btheta^{(t)} | D_n, \tau^{(t)}, \eta^{(t)}, (\sigma^{(t)}_{0})^2) \propto e^{nL(\btheta^{(t)}, D_n)/\tau^{(t)}}\pi_{t}^{\eta^{(t)}/\tau^{(t)}}(\btheta^{(t)})$ for $t = 1, 2, \dots, t_i, \dots, t_f$, \dots, T, where $0 < \eta^{(1)} \leq \eta^{(2)} \leq \dots \leq \eta^{(t_i)}=\eta^{(t_i+1)}=\dots = \eta^{(T)}=1$,  $\tau^{(0)} = \tau^{(1)} = \dots = \tau^{(t_f)} \geq \dots \geq \tau^{(T)}$, and $\pi_t = \lambda N(0, \sigma^{2}_{1}) + (1-\lambda)N(0, (\sigma^{(t)}_{0})^2)$, and $\sigma_{0}^{\text{init}} = \sigma^{(1)}_{0} \geq \sigma^{(2)}_{0} \geq \dots \geq \sigma^{(t_f)}_{0}=\sigma^{(t_f+1)}_{0}=\dots=\sigma^{(T)}_{0}=\sigma^{\text{end}}_{0}$. Denote the resulting model by $\btheta^{(t_f)}$.
   \vspace{2mm}

    \State \textbf{Structure Sparsification:} For each model parameter $i \in \{1, 2,\dots,d\}$, set $\theta_i^{(t_f)} = 1$, if $|\theta_i^{(t_f)}| > \dfrac{\sqrt{2}\sigma_{0}\sigma_{1}}{\sqrt{\sigma_{1}^2 - \sigma_{0}^2}}\sqrt{\log\left(\dfrac{1-\lambda}{\lambda}\dfrac{\sigma_{1}}{\sigma_{0}}\right)}$ and 0 otherwise, and where $\sigma_{0} = \sigma_{0}^{\text{end}}$.
    \vspace{2mm}

    \State \textbf{Input:} training dataset $D_n$, sparsified model $\btheta^{(t_f)}$, number of refining epochs $E_r$, mini-batch size $m_r$.

    \State \textbf{Initialize:} $t=1$, $T_r = \ceil*{E_r n/m_r}$, an optimizer.
    
    \State \textbf{Nonzero-weights Refining:} Refine the nonzero weights of the sparse model $\btheta^{(t_f)}$ by minimizing $L(\bbeta^{(t_f)}, D_n)$.
\end{algorithmic}
\end{algorithm}

In practice, for model pruning, one can utilize a standard optimizer such as Adam or AdamW, rather than employing stochastic gradient MCMC optimizers. A linear scheduler is implemented for $\eta^{(t)}$, $\tau^{(t)}$, and $\sigma_0^{(t)}$ \citep{sun2021sparse, zhang2023sparse}.

\begin{equation}
\label{pa_linear_scheduler}
      \sigma_0^{(t)}, \eta^{(t)}, \tau^{(t)} = \begin{cases} 
      \sigma_0^{\text{init}}, \dfrac{t}{t_i},  \tau^{(0)} & t < t_i \\
      \sigma_0^{\text{end}} + (\sigma_0^{\text{init}} - \sigma_0^{\text{end}})\left(1 - \frac{t-t_i}{t_f-t_i}\right), 1, \tau^{(0)} & t_i \leq t \leq t_f \\
      \sigma_0^{\text{end}}, 1, \dfrac{\tau^{(0)}}{t-t_f} &  t_f < t \leq T
    \end{cases}
\end{equation}

\section{Hyperparameter Sensitivity Analysis}
\label{appendix:hs}

The proposed method, MGPP, comprises six hyperparameters: three originating from the MGP and the other three from the cubic sparsity scheduler. It is important to note that the cubic sparsity scheduler is also employed by the baselines considered in this work. Therefore, when comparing with these recent baselines, we are not introducing additional hyperparameters. In this section, we will specifically address the sensitivity of the three hyperparameters introduced by the MGP.

\begin{itemize}
    \item $\lambda$: MGPP is robust to this hyperparameter, hence we fixed it to $1e-7$ for all experiments.

    \item $\sigma_0^2$: A general guidance for selecting this hyperparameter is to choose a smaller value when given more training samples. We restrict ourselves to the set $\{1e-9, 1e-10\}$.

    \item $\sigma_1^2$: Similarly, a smaller value should be selected when more training samples are available. We restrict ourselves to the set $\{0.1, 0.05\}$.
\end{itemize}

\section{Experiments}
\label{appendix:all_exp}

In this section, we provide detailed information about our experiments, all of which were conducted on A100-80GB GPUs. For all experiments involving MGPP, we set $\lambda$ to $1e-7$, and select the $\sigma_0^2$ from $\{1e-9, 1e-10\}$ and the $\sigma_1^2$ from $\{0.1, 0.05\}$. We choose the learning rates from the set $\{1e-4, 9e-5, 8e-5, 7e-5, 5e-5, 2e-5, 1e-5\}$ and select batch sizes from $\{8, 16, 32, 64\}$. Each experiment adheres to the same number of training epochs as described in \citep{li2023losparse}. For the hyperparameters of the cubic sparsity scheduler, we maintain the same $\Delta t$ value as in \citep{li2023losparse}. Although we largely adhere to their $t_i$ and $t_f$ values, we make necessary adjustments based on our chosen batch size. We use the AdamW optimizer for all experiments.

\subsection{Natural Language Understanding}
\label{appendix:nlu}

Table \ref{table:glue_statistics} provides the dataset statistics of the GLUE benchmark \citep{wang2018glue}, while Table \ref{table:glue_debertav3-base_hyper} details the training hyperparameters for DeBERTaV3\textsubscript{base}, and Table \ref{table:hyperparameters-downstream-bert_base} presents the training hyperparameters for BERT\textsubscript{base}.

\begin{table*}[htbp]
\caption{Summary of the GLUE benchmark.} 
\label{table:glue_statistics}
\begin{center}
\begin{adjustbox}{max width=\textwidth}
\begin{tabular}{ c  l  c  c  c  c  c }
\toprule
\textbf{Corpus} & Task & \#Train & \#Dev & \#Test & \#Label & Metrics \\
\midrule
\multicolumn{7}{c}{Single-Sentence Classification (GLUE)} \\
\midrule
CoLA & Acceptability & 8.5k & 1k  & 1k   & 2 & Matthews corr (Mcc)\\
SST  & Sentiment     & 67k  & 872 & 1.8k & 2 & Accuracy (Acc)\\
\midrule
\multicolumn{7}{c}{Pairwise Text Classification (GLUE)} \\
\midrule
MNLI & NLI         & 393k  & 20k  & 20k    & 3 & Accuracy (Acc) \\
RTE  & NLI         & 2.5k  & 276  & 3k     & 2 & Accuracy (Acc) \\
QQP  & Paraphrase  & 364k  & 40k  & 391k   & 2 & Accuracy/F1 (Acc/F1)\\
MRPC & Paraphrase  & 3.7k  & 408  & 1.7k   & 2 & Accuracy/F1 (Acc/F1)\\
QNLI & QA/NLI      & 108k  & 5.7k & 5.7k   & 2 & Accuracy (Acc) \\
\midrule
\multicolumn{7}{c}{Text Similarity (GLUE)} \\
\midrule
STS-B & Similarity & 7k & 1.5k & 1.4k & 1 & Pearson/Spearman corr (P/S corr)\\
\bottomrule
\end{tabular}
\end{adjustbox}
\end{center}
\end{table*}

\begin{table*}[htbp]
\caption{Hyperparameter setup for MGPP on the GLUE benchmark to prune DeBERTaV3\textsubscript{base}.} 
\label{table:glue_debertav3-base_hyper}
\begin{center}
\begin{adjustbox}{max width=\textwidth}
\begin{tabular}{c c  c c c c c c c c}
\toprule
\textbf{Sparsity} & \textbf{Hyperparameter} & \textbf{MNLI} & \textbf{RTE} & \textbf{QNLI} & \textbf{MRPC} & \textbf{QQP} & \textbf{SST-2} & \textbf{CoLA} & \textbf{STS-B} \\ 
\midrule
\multirow{7}{*}{---}  
    & \#epochs           & 8     & 20    & 10    & 10   & 10    & 6     & 5    & 5     \\
    & Batch size         & 32    & 16    & 64    & 16   & 32    & 32    & 32   & 16    \\
    & $\Delta t$         & 10    & 10    & 10    & 10   & 10    & 10    & 10   & 10  \\
    & $t_i$              & 5500  & 1000  & 1500  & 750  & 10000 & 2000  & 1500 & 1000  \\
    & $t_f$              & 75500 & 2500  & 11500 & 2300 & 85000 & 8000  & 3500 & 3500  \\
    & $\sigma_{0}^2$     & 1e-10 & 1e-10 & 1e-10 & 1e-9 & 1e-10 & 1e-10 & 1e-9 & 1e-9  \\
    & $\sigma_{1}^2$     & 0.05  & 0.1   & 0.1   & 0.1  & 0.1   & 0.1   & 0.1  & 0.1   \\
\midrule
\multirow{1}{*}{80\%} 
    & Learning rate      & 5e-5  & 1e-4  & 8e-5  & 9e-5 & 1e-4  & 7e-5  & 1e-4 & 1e-4 \\
\midrule
\multirow{1}{*}{85\%} 
    & Learning rate      & 5e-5  & 1e-4  & 8e-5  & 1e-4 & 1e-4  & 8e-5  & 1e-4 & 1e-4 \\
\midrule
\multirow{1}{*}{90\%} 
    & Learning rate      & 8e-5  & 1e-4  & 8e-5  & 1e-4 & 1e-4  & 1e-4  & 1e-4 & 1e-4 \\
\bottomrule
\end{tabular}
\end{adjustbox}
\end{center}
\end{table*}

\begin{table*}[htbp]
\caption{Hyperparameter setup for MGPP on the GLUE benchmark to prune BERT\textsubscript{base}.} 
\label{table:hyperparameters-downstream-bert_base}
\begin{center}
\begin{adjustbox}{max width=\textwidth}
\begin{tabular}{c | c c c c c}
\toprule
\textbf{Hyperparameter} & \textbf{MNLI} & \textbf{QQP} & \textbf{QNLI} & \textbf{SQuAD} & \textbf{SST-2} \\ 
\midrule
\multirow{5}{*}{}  
epochs           & 8   & 10  & 10  & 10  & 6 \\
Batch size       & 32  & 32  & 32  & 16  & 32 \\
$\Delta t$       & 100 & 100 & 100 & 100 & 10 \\
$t_i$            & 1 epoch  & 2 epoch & 2 epoch & 2 epoch & 1000 iterations \\
$\sigma_{0}^2$   & 1e-10 & 1e-10 & 1e-9 & 1e-10 & 1e-9 \\
$\sigma_{1}^2$   & 0.05  & 0.05   & 0.05   & 0.05  & 0.1 \\
Learning rate    & \multicolumn{5}{c}{Linearly decay from 5e-5 to 5e-6} \\
$\lambda$        & \multicolumn{5}{c}{$1\times10^{-7}$} \\
\bottomrule
\end{tabular}
\end{adjustbox}
\end{center}
\end{table*}

\subsection{Question Answering}
\label{appendix:qs}

Table \ref{table:squad_debertav3-base_hyper} provides details of the training hyperparameters

\begin{table*}[htbp]
\caption{Hyperparameter setup for MGPP on the SQuAD-v1.1 dataset to prune DeBERTaV3-base.} 
\label{table:squad_debertav3-base_hyper}
\begin{center}
\begin{adjustbox}{max width=\textwidth}
\begin{tabular}{cc c c c c c c c}
\toprule
\textbf{Sparsity} & \#epochs & Batch size & Learning rate & $\Delta t$ & $t_i$ & $t_f$ & $\sigma_0^2$ & $\sigma_1^2$ \\
\midrule
50\% & 10 & 16 & 5e-5 & 10 & 10000  & 35000 & 1e-10 & 0.05 \\
60\% & 10 & 16 & 5e-5 & 10 & 10000  & 35000 & 1e-10 & 0.05 \\
70\% & 10 & 16 & 5e-5 & 10 & 10000  & 35000 & 1e-10 & 0.05 \\
80\% & 10 & 16 & 5e-5 & 10 & 10000  & 35000 & 1e-10 & 0.05 \\
90\% & 10 & 16 & 5e-5 & 10 & 5000   & 40000 & 1e-10 & 0.05 \\
95\%  & 10 & 16 & 5e-5 & 10 & 5000   & 40000 & 1e-10 & 0.05 \\
\bottomrule
\end{tabular}
\end{adjustbox}
\end{center}
\end{table*}

\subsection{Natural Language Generation}
\label{appendix:nlg}

Table \ref{table:nlg_bart-large_hyper} provides details of the training hyperparameters

\begin{table*}[htbp]
\caption{Hyperparameter setup for MGPP on XSum/CNN\_DailyMail datasets to prune BART-large.} 
\label{table:nlg_bart-large_hyper}
\begin{center}
\begin{adjustbox}{max width=\textwidth}
\begin{tabular}{c c c c}
\toprule
\textbf{Sparsity} & \textbf{Hyperparameter} & \textbf{XSum} & \textbf{CNN\_DailyMail}  \\ 
\midrule
\multirow{8}{*}{70\%, 60\%, 50\%}  
    & \#epochs           & 12     & 12      \\
    & Batch size         & 32     & 32      \\
    & $t_i$              & 20000  & 20000   \\
    & $t_f$              & 60000  & 90000   \\
    & $\Delta t$         & 100    & 100   \\
    & $\sigma_{0}^2$     & 1e-10  & 1e-10   \\
    & $\sigma_{1}^2$     & 0.1    & 0.1     \\
    & Learning rate      & 2e-5   & 2e-5     \\
\bottomrule
\end{tabular}
\end{adjustbox}
\end{center}
\end{table*}

\subsection{Upstream Pruning}
\label{appendix:upstream_pruning}

Given the extensive size of the two datasets used in the upstream pruning process, pruning was carried out on 4 V100-32GB GPUs, with each epoch requiring approximately one day to complete. Due to computational resource constraints, we adopted the hyperparameters used for the MNLI dataset from our downstream pruning experiments, with adjustments including lowering $\sigma_1^2$ to 0.01 and setting the pruning frequency to every 200 iterations. For other relevant hyperparameters, we adhered to those utilized by oBERT.

\begin{itemize}
    \item Batch size: 256.
    \item Learning rate: linear decay from 5e-5 to 5e-6.
    \item Maximum sequence length: 512.
    \item Number of epochs for pruning: 3.
\end{itemize}

For the sparse fine-tuning stage, we use the same hyperparameters across all datasets.

\begin{itemize}
    \item Batch size: 32.
    \item Learning rate: linear decay from 2e-5 to 0.
    \item Maximum sequence length: 512.
    \item Number of epochs: 8.
\end{itemize}

\subsection{Ablation Study}
\label{appendix:as}

For the $L_2$ ablation variant, we set the weight decay coefficient to $1e-2$, chosen from the set $\{0.1, 1e-2, 1e-3, 1e-4, 1e-5\}$. Additional hyperparameters are detailed in Table \ref{table:l2_ablation}.

\begin{table*}[htbp]
\caption{Hyperparameter setup for $L_2$ on the GLUE benchmark to prune DeBERTaV3-base.} 
\label{table:l2_ablation}
\begin{center}
\begin{adjustbox}{max width=\textwidth}
\begin{tabular}{c c c c c }
\toprule
\textbf{Sparsity} & \textbf{Hyperparameter} & \textbf{MNLI} & \textbf{MRPC} & \textbf{SST-2} \\ 
\midrule
\multirow{6}{*}{80\%, 85\%, 90\%}  
    & \#epochs           & 8     & 10    & 6     \\
    & Batch size         & 32    & 16    & 32    \\
    & $\Delta t$         & 10    & 10    & 10    \\
    & $t_i$              & 5500  & 1000  & 1500  \\
    & $t_f$              & 75500 & 2500  & 11500 \\
    & Learning rate      & 5e-5  & 9e-5  & 5e-5  \\
\bottomrule
\end{tabular}
\end{adjustbox}
\end{center}
\end{table*}

For the $PA$ ablation variant, we conducted an extensive search for the optimal hyperparameters. The specific hyperparameters are detailed in Table \ref{table:pa_ablation}.

\begin{table*}[!ht]
\caption{Hyperparameter setup for PA on the GLUE benchmark to prune DeBERTaV3-base.} 
\label{table:pa_ablation}
\begin{center}
\begin{adjustbox}{max width=\textwidth}
\begin{tabular}{c c  c c c }
\toprule
\textbf{Sparsity} & \textbf{Hyperparameter} & \textbf{MNLI} & \textbf{MRPC} & \textbf{SST-2} \\ 
\midrule
\multirow{8}{*}{---}  
    & $E$                 & 7     & 9     & 5     \\
    & $m$                 & 32    & 16    & 32    \\
    & $E_r$               & 1     & 1     & 1     \\
    & $m_r$               & 32    & 16    & 32    \\
    & $\tau^{(0)}$        & 1     & 1      & 1   \\
    & $\lambda$           & 1e-7  & 1e-7   & 1e-7  \\
    & $\sigma_{1}^2$      & 0.05  & 0.1    & 0.5   \\
    & learning rate       & 2e-5  & 5e-5   & 2e-5   \\
\midrule
\multirow{4}{*}{80\%} 
    & $t_i$              & 5000  & 1000  & 1500  \\
    & $t_f$              & 80000 & 3000  & 12000 \\
    & $(\sigma_{0}^{\text{init}})^2$ & 1e-4 & 1e-4 & 1e-4 \\
    & $(\sigma_{0}^{\text{end}})^2$  & 1e-5 & 1e-5  & 1e-5 \\
\midrule
\multirow{4}{*}{85\%} 
    & $t_i$              & 5000  & 1000  & 1500  \\
    & $t_f$              & 80000 & 2500  & 12000 \\
    & $(\sigma_{0}^{\text{init}})^2$ & 1.2e-4 & 1.2e-4 & 1.2e-4 \\
    & $(\sigma_{0}^{\text{end}})^2$  & 2e-5 & 2e-5  & 2e-5 \\
\midrule
\multirow{4}{*}{90\%}
    & $t_i$              & 5000  & 1000  & 1500  \\
    & $t_f$              & 80000 & 3000  & 12000 \\
    & $(\sigma_{0}^{\text{init}})^2$ & 1.4e-4 & 1.4e-4 & 1.4e-4 \\
    & $(\sigma_{0}^{\text{end}})^2$  & 3e-5   & 3e-5   & 3e-5 \\
\bottomrule
\end{tabular}
\end{adjustbox}
\end{center}
\end{table*}


 \bibliography{reference}
 \bibliographystyle{apalike}

\end{document}